\documentclass[11pt]{article}

\usepackage[preprint]{acl}

\usepackage{times}
\usepackage{latexsym}
\usepackage[T1]{fontenc}
\usepackage[utf8]{inputenc}
\usepackage{microtype}
\usepackage{inconsolata}
\usepackage{graphicx}
\usepackage{booktabs}
\usepackage{amsmath}
\usepackage{multirow}
\usepackage{enumitem}
\usepackage{pgfplots}
\pgfplotsset{compat=1.18}
\usepackage{fontawesome5}

\newcommand{\yr}[1]{(\citeyear{#1})}

\title{\raisebox{-8pt}{\includegraphics[height=2em]{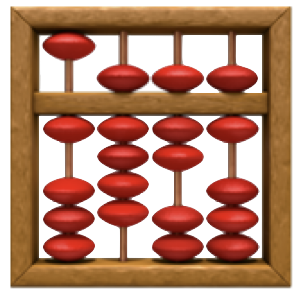}} ABAQUS: The Quantitative Language Resource Index}

\author{Jana Massoud \qquad\qquad Sina Ahmadi \\
  Department of Computational Linguistics, University of Zurich \\
  \texttt{\{jana.massoud, sina.ahmadi\}@uzh.ch} \\}

\begin{document}
\maketitle

\begin{abstract}
Despite the widespread invocation of ``low-resource'' in natural language processing (NLP), the term remains a vague and inconsistent measure of a language's actual standing, obscuring the vast asymmetries in available technological and linguistic artifacts. This paper introduces ABAQUS, a quantitative resource index that evaluates 8,168 languages documented in Glottolog across six dimensions: corpora, models, benchmarks, grammars, dictionaries, and parallel data. By combining these dimensions into a continuous, reconfigurable composite score, ABAQUS moves beyond coarse binary classifications to offer a fine-grained, transparent measure of language resource availability. Our findings reveal a heavily right-skewed landscape where nearly three-quarters of languages have effectively no presence in contemporary NLP artifacts, while highlighting languages whose practical resource availability exceeds their perceived standing. Validating the index against downstream performance on five tasks, we find that individual dimensions are strongly task-diagnostic, while the composite best serves as a landscape-level summary rather than a per-task predictor. ABAQUS provides a robust quantitative foundation for situating languages within the current technological ecosystem and identifying critical gaps for future development. 
\vspace{-0.1cm}
\begin{center}
\raisebox{-0.12cm}{\includegraphics[height=0.45cm]{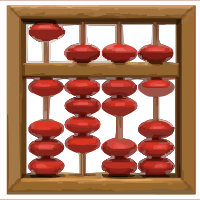}}~%
\href{https://zurichnlp.github.io/ABAQUS/}{\textbf{zurichnlp.github.io/ABAQUS}}
\end{center}
\end{abstract}

\section{Introduction}

\begin{figure}[t]
    \centering
    \includegraphics[width=\columnwidth]{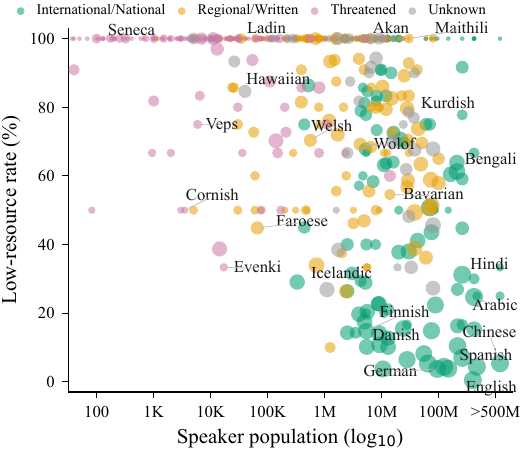}
    \caption{Distribution of language varieties labeled as ``low-resource'' or similar terms in the titles and abstracts of papers indexed on the ACL Anthology (1963--2026). While international and national languages are less often described as ``low-resourced'' less, there is no consistent definition of the term based on population or vitality. Speaker-population figures are drawn from GlotWeb~\citep{sefat2026glotweb} and institutional vitality from EGIDS in Glottolog~\citep{glottolog5.3}.} 
    \label{fig:lowresource}
    \vspace*{-0.5cm}
\end{figure}

Whether used to describe the authentic scarcity of resources available for a language or to signal the novelty of a task being addressed for the first time, ``low-resource'' has become one of the most frequently invoked labels in natural language processing (NLP). Yet the notion is far from uniform across languages. Basque being referred to as ``low-resource''~\cite{DBLP:conf/emnlp/SainzPELAGZARAAS25} is fundamentally different from what the term means for low-resourced Arabic dialects~\citep{qaddoumi-etal-2026-learning}, Hawrami~\citep{ahmadi-etal-2025-parme}, or Wixárika~\citep{gagnier-kirubakaran-2026-deer}. Because virtually any language can be considered data-scarce depending on the task, domain, or modality, ``low-resource'' has become a vague and often inconsistent measure of a language's actual standing in NLP. Of the 80{,}314 papers indexed on the ACL Anthology (1963--July 2026), 22{,}504 (28\%) name at least one language in their title or abstract~\citep[cf.][]{bender-friedman-2018-data}; among those, 4{,}789 (6\% of all papers) describe a language as ``low-resource'' or a similar term.\footnote{Language mentions and low-resource framing were extracted from paper titles and abstracts using Gemini 3.5 Flash-Lite.} The label is applied to 1{,}073 distinct languages ranging from a few hundred to over a billion speakers, with no consistent grounding in population or vitality, as shown in Figure~\ref{fig:lowresource}.

The stakes of this ambiguity have grown sharper as large language models (LLMs) extend nominal multilingual coverage to hundreds and thousands of languages, while the gap between claimed and actual capability remains largely unmeasured. Resource availability is neither a single quantity nor a static property: it varies across the artifacts a language possesses, the tasks those artifacts support, and the communities whose priorities and participation shape what counts as a resource in the first place~\citep{DBLP:conf/acl/Bird22, liu-etal-2022-always}. Treating this landscape with coarse binary or few-class labels obscures precisely the asymmetries that matter most, and risks reinforcing a trajectory in which multilingual claims outpace substantive language coverage~\citep{bird-2026-big}. As such, the absence of a fine-grained, transparent, and reconfigurable measure of resource availability is not a methodological inconvenience but a structural gap with concrete consequences for the communities whose visibility in NLP depends on how their status is characterized~\citep{DBLP:conf/chi/NigatuCC24}.

Existing efforts to quantify resource availability have laid essential groundwork but stop short of the granularity and flexibility the current landscape demands. Class-based taxonomies provide a common vocabulary but collapse meaningful within-class variation~\citep{joshi-etal-2020-state}, while demand- and utility-weighted approaches capture socio-technical dimensions at the cost of task-specific configurability~\citep{blasi-etal-2022-systematic}, and web-presence surveys expand the empirical basis without addressing the heterogeneity of resource types that different tasks require~\cite{sefat2026glotweb}. This paper introduces ABAQUS, a quantitative index that scores 8,168 languages documented in Glottolog along six resource dimensions, namely corpora, models, benchmarks, grammars, dictionaries, and parallel data, and combines them into a continuous composite score. 
We present our contributions across three axes:
\begin{itemize}[noitemsep,nolistsep]
    \item \textbf{Data:} A quantitative assessment of resource availability spanning the world's languages.
    \item \textbf{Task:} An empirical validation across five downstream tasks showing which resource dimensions predict which tasks.
    \item \textbf{Framework:} A framework that enables practitioners and community stakeholders to situate a given language against the current NLP landscape and to surface asymmetries.
\end{itemize}

\section{What Is a Low-Resource Language?}
\label{sec:what-is-lrl}

A substantial body of work defines low-resource in terms of a language's linguistic description, or lack thereof: a language is low-resource when little is known of its structure and few artifacts archive that knowledge \citep{nigatu-etal-2024-zenos}. Beyond documentation, such resources also shape NLP methodology. \citet{knowles-littell-2022-translation}, for instance, show that morphologically rich languages require different approaches to lexical variety and out-of-vocabulary tokens, which grammars and descriptions enable. However, the existence alone of linguistic documentation is not sufficient. \citet{singh-2008-natural} notes that even well-studied languages can be low-resource if their descriptions are not machine-readable, since many linguistic descriptions exist only as scanned, irregularly formatted text \citep{zhang-etal-2024-hire}. This scarcity of usable data is the starting point for most low-resource NLP surveys \citep{hedderich-etal-2021-survey}.

A second line of work classifies low-resource languages (LRLs) in social terms, describing their speaker communities as being in ``survival mode''. Endangered languages, defined by \mbox{\citet{moseley2010atlas}} as those no longer being passed down to younger generations, are the clearest case. However, \citet{DBLP:journals/corr/abs-2103-09567} cautions against treating endangered languages as a homogeneous subset of LRLs as they vary in urgency, require different revival strategies, and face challenges that low-resource NLP methods alone cannot address. A related axis is socio-political standing. \citet{nigatu-etal-2024-zenos} argue that a lack of social recognition as well as limited participation in government and education is itself indicative of an LRL. This standing is not static: \citet{mehta2020gondi} show that NLP interest in a socially disadvantaged language can be triggered by politics, such as a government-supported language education. \citet{singh-2008-natural} similarly identifies social and political recognition as a central barrier to progress.

This multifaceted view is reflected in the field's own shifting terminology \citep{cieri-etal-2016-selection}. \citet{singh-2008-natural} groups these languages as L-languages, alluding to their common description as less studied, less resourced, less computerized, and less privileged. \citet{bird-2020-decolonising} critiques such deficit-centered framing for reproducing the very hierarchies it describes and proposes ``local languages'' instead, foregrounding each case's ``local lifeworld and culture area'' \citep{DBLP:conf/acl/Bird22}.

\begin{table*}[!t]
\centering
\small
\renewcommand{\arraystretch}{1.2}
\begin{tabular}{lccl}
\toprule
\textbf{Approach} & \textbf{Languages} & \textbf{Result per language} & \textbf{Resource signals measured} \\
\midrule
\citet{joshi-etal-2020-state}
  & $\sim$2,500 & one of 6 ordered classes
  & labeled + unlabeled data volume \\
\citet{blasi-etal-2022-systematic}
  & global & per-task utility score
  & task performance, speaker demographics \\
\citet{ranathunga-de-silva-2022-languages}
  & 6{,}420 & coverage across 4 axes
  & data, research output, web platforms, \\
  & & & pretrained-model inclusion \\
\citet{gaspari-etal-2022-introducing}
  & Europe only & weighted score (dashboard)
  & catalogued tools \& resources, \\
  & & & socio-economic context \\
\citet{coleman-etal-2026-ran}
  & extensible & multi-part notation string
  & speakers, monolingual data, \\
  & & & parallel data per partner \\
\midrule
\textbf{ABAQUS (ours)}
  & \textbf{8{,}168} & \textbf{composite score, 0--100}
  & \textbf{corpora, models, benchmarks,} \\
  & & & \textbf{grammars, dictionaries, parallel data} \\
\bottomrule
\end{tabular}
\caption{Approaches to quantifying language resource level in NLP, compared by language coverage, the form of their per-language output, and the resource signals they draw on. ABAQUS is the only global, continuous index spanning six resource dimensions across both NLP artifacts and linguistic documentation. Position papers that motivate quantification without performing it are not included in the table.}
\label{tab:prior-quant}
\vspace{-0.5cm}
\end{table*}

\subsection{Measuring Language Resource Level}
\label{sec:measuring}

Efforts to make resource level concrete range from catalog-based counting to performance-based estimation. The most direct approach reads resource availability off curated repositories and catalogs. Bibliographic databases such as Glottolog and its Langdoc reference collection \citep{nordhoff2011glottolog} list the grammars and descriptions available for each language. \citet{zhang-etal-2024-hire} use these statistics to assess the availability of grammar books and dictionaries. Others rely on infrastructure catalogs: \citet{platzgummer2026} use the CLARIN Virtual Language Observatory\footnote{\url{https://vlo.clarin.eu}} to assess the resource landscape for German varieties. Such catalogs, along with repositories like the Linguistic Data Consortium,\footnote{\url{https://www.ldc.upenn.edu}} offer reproducible counts but inherit their own coverage biases. A related, coarser convention defines the boundary operationally: \citet{caswell-etal-2025-smol} treat as low-resource any language beyond the roughly 100 supported by mainstream crawls and translation providers like Google Translate.

A second approach converts such counts into a discrete taxonomy. \citet{joshi-etal-2020-state} sort languages into a six-class taxonomy based on the volume of labeled and unlabeled data available for each language. \citet{kreutzer-etal-2022-quality} and \citet{lau-etal-2025-data} complicate any purely volumetric measure by auditing the quality of multilingual corpora, finding that lower-resource languages suffer disproportionately from low-quality data, so that raw counts overstate their true standing.

\begin{table}[htbp!]
\centering
\small
\setlength{\tabcolsep}{3pt}
\renewcommand{\arraystretch}{0.85}
\begin{tabular}{@{}clcr@{}}
\toprule
& \textbf{Resource} & \textbf{Type / Task} & \textbf{Langs} \\
\midrule
\multirow{9}{*}{\rotatebox[origin=c]{90}{\textbf{Corpora}}}
  & FineWeb2~\yr{penedo2025fineweb}       & Web crawl            & 1871 \\
  & MADLAD-400~\yr{kudugunta2023madlad} & Crawl \& parallel & 417 \\
  & HPLT v3.0~\yr{oepen2025hplt30largescalemultilingual} & Web crawl            & 198 \\
  & NLLB~\yr{DBLP:journals/nature/Team24}      & Parallel \& bitext   & 163 \\
  & CC-100~\yr{wenzek-etal-2020-cc100}                   & Web crawl            & 115 \\
  & mC4~\yr{raffel2023c4}                                & Web crawl            & 107 \\
  & WikiMatrix~\yr{schwenk-etal-2021-wikimatrix} & Parallel Wiki     & 59  \\
  & OpenSubtitles~\yr{lison-tiedemann-2016-opensubtitles2016} & Parallel subs   & 58  \\
  & ROOTS~\yr{lauren_on2022the}                        & Curated mixture      & 46  \\
\cmidrule(l){2-4}
\multirow{11}{*}{\rotatebox[origin=c]{90}{\textbf{Models}}}
  & Apertus~\yr{swissai2025apertus}                      & 8B / 70B             & 1849 \\
  & mmBERT~\yr{marone2026mmbert} & 140M / 307M         & 1833 \\
  & Omnilingual MT~\yr{omnilingualmtteam2026omnilingualmtmachinetranslation} & 300M--7B & 1681 \\
  & Glot500-m~\yr{ImaniGooghari_2023}                    & 126M                 & 534 \\
  & NLLB~\yr{DBLP:journals/nature/Team24}      & 600M--54.5B          & 204 \\
  & PaLM~\yr{3648699.3648939}   & 8B--540B             & 120 \\
  & XGLM~\yr{lin-etal-2022-shot} & 125M--6.7B           & 113 \\
  & mT5~\yr{xue-etal-2021-mt5}                           & 300M--13B            & 107 \\
  & XLM-R~\yr{conneau-etal-2020-unsupervised} & 270M / 550M & 99 \\
  & BLOOM~\yr{JMLR:v25:23-0581} & 176B      & 46  \\
  & PolyLM~\yr{wei2023polylmopensourcepolyglot}          & 1.7B / 13B           & 18  \\
\cmidrule(l){2-4}
\multirow{9}{*}{\rotatebox[origin=c]{90}{\textbf{Benchmarks}}}
  & Flores-200~\yr{gayol2021flores} & Translation          & 204 \\
  & Global PIQA~\yr{chang2026globalpiqaevaluatingcommonsense} & Commonsense  & 115 \\
  & ALM-bench~\yr{vayani2025alm} & Cultural VQA         & 100 \\
  & XTREME~\yr{hu2020xtrememassivelymultilingualmultitask} & NLI, POS, NER & 40  \\
  & DialectBench~\yr{faisal2024dialectbench} & Multi-task (10)           & 38  \\
  & MMLU-ProX~\yr{xuan-etal-2025-mmlu} & MCQ           & 29  \\
  & BenchMAX~\yr{huang-etal-2025-benchmax} & MT, QA  & 17  \\
  & MAPS~\yr{hofman-etal-2026-maps}   & Code, math           & 12  \\
  & MultiHal~\yr{lavrinovics2025multihal}   & Hallucination QA   & 6  \\
\cmidrule(l){2-4}
\multirow{9}{*}{\rotatebox[origin=c]{90}{\textbf{Parallel Corpora}}}
  & translatewiki~(v2025-01-01)  & 203{,}142 bitexts & 666 \\
  & Tatoeba~(v2023-04-12)        & 4{,}344 bitexts   & 397 \\
  & wikimedia~(v20230407)        & 3{,}656 bitexts   & 322 \\
  & QED~(v2.0a)                  & 9{,}291 bitexts   & 225 \\
  & GNOME~(v1)                   & 12{,}822 bitexts  & 187 \\
  & (Multi)CCAligned~(v1.1\textbackslash v1) & 6{,}155 bitexts       & 113 \\
  & TED2020~(v1)                 & 4{,}796 bitexts   & 108 \\
  & bible-uedin~(v1)             & 5{,}148 bitexts   & 102 \\
  & KDE4~(v2)                    & 4{,}099 bitexts   & 92  \\
\cmidrule(l){2-4}
\multirow{2}{*}{\rotatebox[origin=c]{90}{\textbf{Doc.}}}
  & Grammars~\yr{glottolog5.3}    & $\geq$1 grammar    & 2964 \\
  & Dictionaries~\yr{glottolog5.3} & $\geq$1 dictionary & 2916 \\
\bottomrule
\end{tabular}
\caption{Resources included in ABAQUS, grouped by dimension. Grammar and dictionary counts are from Glottolog \citep{glottolog5.3, nordhoff2011glottolog} and; the figures shown are the number of languages with at least one such resource. parallel corpora are accessed through OPUS \citep{tiedemann-2012-parallel}.}
\label{tab:resources}
\end{table}

A third line abandons class labels for continuous scores. \citet{blasi-etal-2022-systematic} argue that raw data volume alone is insufficient, and instead estimate the global utility of language technologies from published performance, integrating demographic characteristics such as speaker-group population. \citet{ranathunga-de-silva-2022-languages} broaden the set of signals, cross-referencing data availability, research volume, web-platform presence, and inclusion in pretrained multilingual models against a speaker-population-and-vitality categorization. In a European context, the Digital Language Equality metric \citep{gaspari-etal-2022-introducing, grutzner-zahn-rehm-2022-introducing} combines technological factors, drawn from the resources and tools catalogued in the European Language Grid \citep{rehm-etal-2020-european-language}, with contextual socio-economic factors into a single weighted score per language. Most recently, \citet{coleman-etal-2026-ran} propose Resource Abundance Notation (RAN), a compact multidimensional notation summarizing a language's profile as order-of-magnitude counts of speakers, monolingual sentences, and parallel data, designed primarily as a communication tool rather than a single ranking. 

Table~\ref{tab:prior-quant} situates these approaches along three axes: how many languages they reach, whether their output is a class, a score, or a notation, and which resource signals they draw on.

\section{ABAQUS}
\label{sec:abaqus}
We introduce ABAQUS, an index that quantifies the resource availability of every language documented in Glottolog~\citep{glottolog5.3} on a continuous 0--100 scale. Each language receives a composite score which combines six resource dimensions: corpus, model, and benchmark coverage, together with grammar, dictionary, and parallel data availability.\footnote{Sources are matched to Glottolog languages by ISO~639-3 code, after removing script suffixes (\texttt{afr\_Latn} $\rightarrow$ \texttt{afr}).} 

\subsection{Resources}

\paragraph{Corpora} Large multilingual corpora are the raw material of modern NLP, and a language's presence or absence in them is often the first thing that marks it as low-resource. To index this dimension, we collect the language composition of nine multilingual corpora. Reported inclusion in corpora also requires that language identification was successfully run and returned a positive match.\footnote{For some large web-scraped corpora, this is not the case, leading to their exclusion from the index.} This dimension therefore functions as a gating condition rather than a measure of volume.

\paragraph{Models} While the availability of data enables model training, model coverage demonstrates the availability of ready-to-use tools. Model coverage information is collected for eleven multilingual language models.\footnote{Only models which disclose language coverage information were included.} This dimension measures claimed coverage rather than verified performance: a language listed as `within the ability of a model' reflects inclusion in training data, and not quantity of training data nor model performance. Whether a claimed capability is verifiable is addressed by the benchmarks dimension.

\paragraph{Benchmarks} Data availability and model coverage alone do not ensure adequate performance. This index uses language coverage information for nine multilingual benchmarks to measure performance verifiability across languages. A language's absence from benchmarks means its claim under any model or corpus is fundamentally unverifiable. Benchmarks are also designed for specific tasks, so a language's inclusion in one does not guarantee its evaluation for all downstream tasks. 

\paragraph{Grammars and Dictionaries} Where large corpora are unavailable, many approaches leverage data from dictionaries and grammars \citep{zevallos-etal-2024-tema}. We therefore count, for every language, the
grammars and dictionaries listed in Glottolog's bibliographic references.\footnote{This section of Glottolog is available on the \href{https://glottolog.org/langdoc}{`references'} page as well as under each individual language page.} This bibliographic collection is treated as a comparable proxy measure of how many of these resources are available across languages.

\paragraph{Parallel Data} Many downstream NLP tasks require annotated data. Machine translation (MT), for example, requires parallel corpora for training. To account for tasks such as these, the availability of parallel data is evaluated from OPUS \citep{tiedemann-2012-parallel}, a collection of open parallel corpora. For this, metadata from nine corpora were consulted. To avoid duplicate counts, only the newest version of each corpus was used.

Each language $i$ is scored on three signals: how many parallel corpora include it ($n_i$), how many distinct languages it is paired with across all corpora ($p_i$), and the total number of parallel sentences available for it across all corpora ($s_i$). A weighted score is computed as:
\begin{equation}
P_i = 100 \cdot \left( w_n\,\hat{n}_i + w_p\,\hat{p}_i + w_s\,\hat{s}_i \right)
\end{equation}
\noindent where $w_n=0.2$, $w_p=0.4$, and $w_s=0.4$. Each raw count ${x}_i \in \{n_i, p_i, s_i\}$ is log-normalized over all languages $j$ in the dataset before being combined:
\begin{equation}
\hat{x}_i = \frac{\ln(1 + x_i) - \min_{j}\,\ln(1 + x_j)}{\max_{j}\,\ln(1 + x_j) - \min_{j}\,\ln(1 + x_j)}
\end{equation}

\subsection{Composite Index Score}
For every language, a composite score is computed combining the weighted value of each resource dimension. For corpus (C), model (M), and benchmark (B) coverage, maximum scores represent the number of sources included. For grammars (G) and dictionaries (D), artificial maxima (30 and 20, respectively) were selected in order to cap values.\footnote{For some languages, hundreds of grammars and dictionaries are listed. The artificial maximum stops the inflation of scores and ensures that genuinely high-resource languages do not receive comparably low scores.} Parallel data (P) coverage scores (already normalized on a 0--100 scale) are rescaled linearly. The composite resource score $S \in [0, 100]$ is computed as:
\begin{equation}
S = 100\cdot\frac{\sum_{r \in R} w_r \cdot n_r}{\sum_{r \in R} w_r}
\end{equation}
where $R = \{\text{G, D, C, M, B, P}\}$ is the set of resource dimensions, $w_r$ is the assigned weight for dimension $r$, and $n_r \in [0,1]$ is its normalized score:
\begin{equation}
n_r =
\begin{cases}
\min\!\left(1,\, \dfrac{\ln(1 + c_r)}{\ln(1 + M_r)}\right) & r \in R \setminus \{\text{P}\} \\[2mm]
\min\!\left(1,\, \dfrac{c_r}{100}\right) & r = \text{P}
\end{cases}
\end{equation}
where $c_r$ is the raw count for dimension $r$ (for parallel, $c_r=P_i$, the pre-normalized 0--100 score), and $M_r$ is the artificial maximum for the log-normalized dimensions. Log-normalization compresses the count scale, and the $\min(1,\cdot)$ cap ensures no dimension exceeds its maximum contribution.

\begin{table}[t]
\centering
\renewcommand{\arraystretch}{1}
\resizebox{1\columnwidth}{!}{%
\begin{tabular}{lcc}
\toprule
\textbf{Dimension} & \textbf{Weight ($w_r$)} & \textbf{Share of Composite} \\
\midrule
Models        & 6 & 25.0\% \\
Corpora       & 5 & 20.8\% \\
Benchmarks    & 5 & 20.8\% \\
Parallel Data & 4 & 16.7\% \\
Grammars      & 2 & 8.3\%  \\
Dictionaries  & 2 & 8.3\%  \\
\midrule
\textbf{Total} & \textbf{24} & \textbf{100\%} \\
\bottomrule
\end{tabular}
}
\caption{Default weights assigned to each resource dimension in the composite score, sorted by contribution.}
\label{tab:weights}
\end{table}

\begin{table}[t]
    \centering
    \renewcommand{\arraystretch}{1}
    \resizebox{1\columnwidth}{!}{%
    \begin{tabular}{lcccccc}
    \toprule
    \textbf{Task} & \textbf{G} & \textbf{D} & \textbf{C} & \textbf{M} & \textbf{B} & \textbf{P} \\
    \midrule
    MT & 0.1 & 0.3 & 3.0 & 1.9 & 2.8 & 1.8 \\
    BR    & 0.0 & 0.2 & 0.3 & 3.3 & 4.9 & 1.3 \\
    NLI                 & 0.0 & 0.2 & 0.0 & 9.8 & 0.0 & 0.0 \\
    QA  & 0.0 & 0.0 & 0.8 & 3.0 & 5.9 & 0.3 \\
    POS tagging         & 2.8 & 0.0 & 0.0 & 1.3 & 0.0 & 6.0 \\
    \bottomrule
    \end{tabular}
    }
    \caption{Task-specific dimension weights learned via cross-validated non-negative least squares against downstream performance. Weights are scaled to sum to~10. Near-zero weights indicate a dimension is redundant given the others for that task, not that it is irrelevant.}
    \label{tab:weights-learned}
\end{table}

The default weight configuration is provided in Table~\ref{tab:weights}; reported results in this paper are calculated on this basis, representing a general overview of a language's NLP presence.\footnote{Digital resources receive 83\% of the weight, since the index targets readiness for language technology. Among them, models and benchmarks weigh most, as they show a language can be processed and evaluated. Appendix~\ref{app:sensitivity} shows the results are robust to these choices.} In addition to this, task-specific configurations are derived empirically. For five downstream tasks, per-dimension resource scores are correlated against observed model performance to learn dimension weights that best predict it; more details on evaluating performance are provided in Appendix~\ref{sec:validation}). Table~\ref{tab:weights-learned} reports the resulting configurations. Traditional documentation receives little to no weight overall, with POS tagging as the only task to assign substantial weight to grammars (2.8). MT distributes its weight across corpora, benchmarks, models, and parallel data. Bitext retrieval (BR) and QA are weighted mainly towards benchmark and model coverage. NLI places nearly all its weight on model coverage. These configurations support the intuition that different tasks depend on different resource types, and ground it in measured performance rather than assignment.

\section{Analysis}

\subsection{Resource Distribution}
\begin{figure}[t]
\centering
\includegraphics[width=\columnwidth]{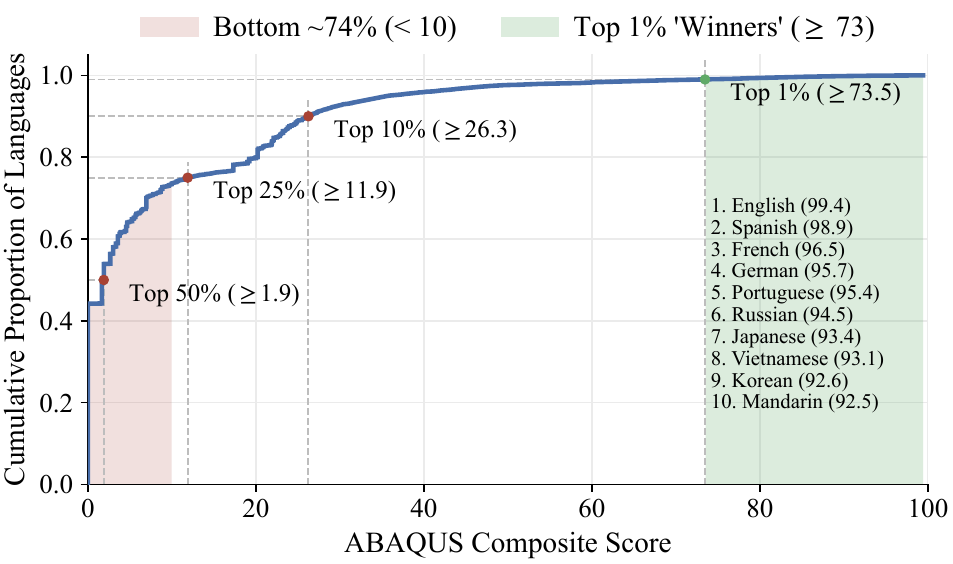}
\caption{Cumulative distribution function of ABAQUS composite scores across all 8,168 languages. The red shaded region highlights the severe long tail of the NLP resource distribution, where nearly 74\% of languages receive a score below 10. Score thresholds for key percentiles are provided in parentheses. The inset lists the 10 highest-scoring languages globally.}
\label{fig:cdf-scores}
\end{figure}

As illustrated in the cumulative distribution function (Figure~\ref{fig:cdf-scores}), the resource distribution across the world's languages is heavily right-skewed. \textbf{Less than 1\% of languages receive a composite score above 75, and only 2.4\% receive one above 50}. The shaded region of the plot demonstrates that 73.6\% of languages (6,012 languages) score below 10, more than half of which are assigned a score of 0. The mean composite score across all languages is 8.77; excluding zero-scoring languages, it rises only to 15.71. As such, resource scarcity is not confined to a residual tail consisting of extreme zero-scoring cases, but is the default state for the majority of documented languages.

\begin{figure*}[t]
\centering
\includegraphics[width=\textwidth]{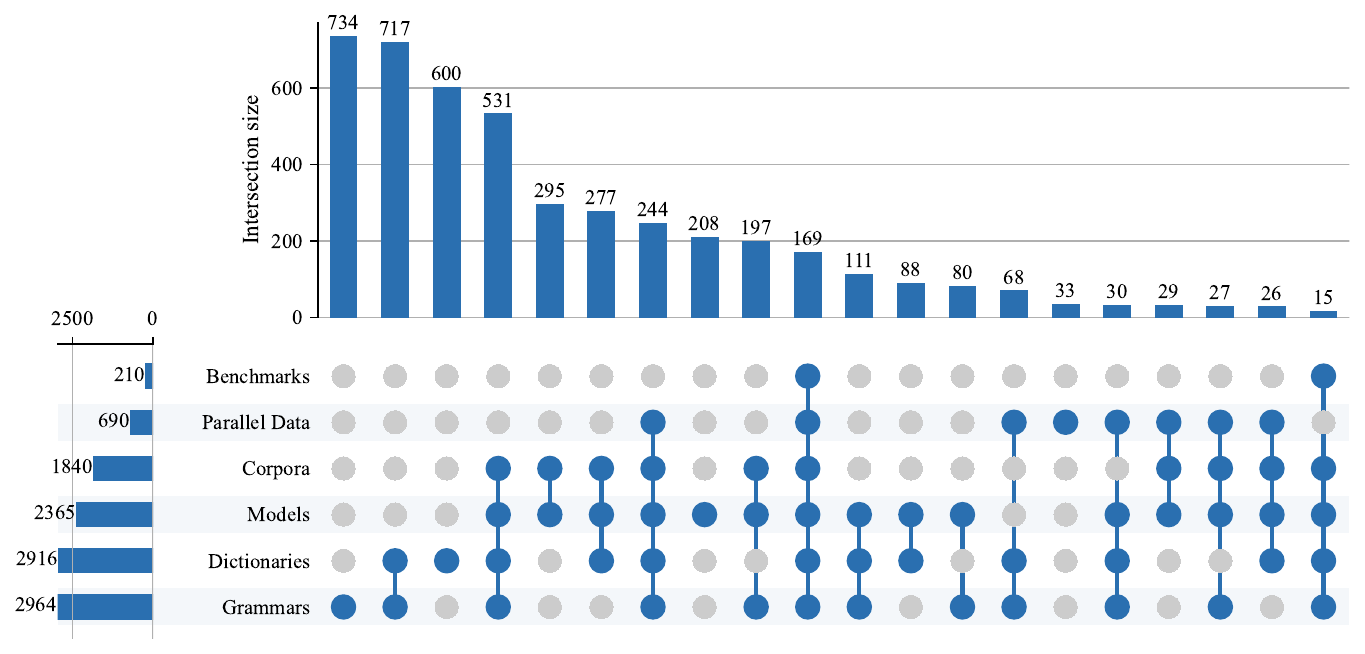}
\caption{UpSet plot detailing the intersections of resource availability across the six ABAQUS dimensions for active languages (excluding the 3,609 zero-resource entries). This visualizes the documentation-to-digitization gap, demonstrating that while thousands of languages possess traditional linguistic resources (grammars and dictionaries), they are systematically excluded from digital artifacts and evaluative benchmarks.}
\label{fig:upset-resources}
\end{figure*}

The top 20 highest-resource languages feature the ``Winners'' defined by \citet{joshi-etal-2020-state}: English, Spanish, French, German, Japanese, Arabic, and Mandarin. 
For the 1{,}882 languages in both resources, ABAQUS preserves the class hierarchy. The median scores rise from 7.7 (class 0) to 95.7 (class 5), and in 96.2\% of cross-class pairs the higher class scores higher. However, the classes explain only 61\% of the ABAQUS score variance. Class 0 alone spans 0--60 with 12.0\% zero-scoring languages, 40.7\% only documentation, and 47.3\% at least one digital resource.

Examining per-dimension contributions reveals severe structural bottlenecks in the NLP pipeline. As visualized in the UpSet plot~\citep{lex2014upset} in Figure~\ref{fig:upset-resources}, which maps resource combinations across the active subset of evaluated languages, the dimensions with the widest coverage are traditional linguistic resources: 2,964 languages (36.3\%) possess at least one grammar, closely followed by dictionaries at 2,916 languages (35.7\%). However, as we move toward the artifacts required for modern machine learning, coverage collapses. Benchmark inclusion is the most severe bottleneck, with only 210 languages (2.6\%) receiving non-zero scores. The intersections in Figure~\ref{fig:upset-resources} highlight this gap: a vast number of languages possess traditional documentation but remain entirely disconnected from the digital datasets, models, and evaluative benchmarks that define contemporary NLP.

\begin{table}[t]
\centering
\resizebox{0.8\columnwidth}{!}{%
\begin{tabular}{@{}lcccccc@{}}
\toprule
\textbf{Task} & \textbf{G} & \textbf{D} & \textbf{C} & \textbf{M} & \textbf{B} & \textbf{P} \\
\midrule
MT & 0.21 & 0.22 & \textbf{0.49} & 0.37 & 0.48 & 0.46 \\
BR  & 0.18 & 0.17 & 0.31 & 0.27 & \textbf{0.47} & 0.41 \\
NLI               & 0.18 & 0.17 & 0.27 & 0.21 & \textbf{0.36} & 0.34 \\
QA  & $-$0.16 & $-$0.22 & 0.60 & 0.57 & 0.67 & \textbf{0.69} \\
POS       & \multicolumn{6}{c}{\textit{inconclusive}} \\
\bottomrule
\end{tabular}
}
\caption{Spearman's rank correlation ($\rho$) between each resource dimension and downstream task performance. The strongest correlate per task is in bold. 95\% confidence intervals (Fisher $z$) are $\pm$0.05--0.07 for MT, BR and NLI and $\pm$0.10--0.19 for QA.}
\label{tab:per-dim-corr}
\end{table}

\subsection{What the Dimensions Predict}
\label{sec:validation-body}

The composite score summarizes overall resource presence, but does each dimension reflect real technological disparity, and does the composite predict performance on specific tasks? To find out, we correlate the per-dimension scores against observed downstream performance on five tasks spanning generation, retrieval, inference, comprehension, and sequence labeling: MT and BR over 807 languages, NLI over 833, reading comprehension over 112, and POS tagging over 71. Full protocol, models, and metrics are given in Appendix~\ref{sec:validation}.

\paragraph{Each task is predicted by a specific dimension.} Table~\ref{tab:per-dim-corr} reports the per-dimension Spearman correlations with performance. MT correlates most with corpus ($\rho=0.49$), benchmark ($\rho=0.48$), and parallel data ($\rho=0.46$), consistent with generation depending on monolingual and aligned text; BR and NLI with benchmark coverage ($\rho=0.47$, $0.36$); and reading comprehension with parallel and benchmark coverage ($\rho=0.69$, $0.67$). Reading comprehension shows weak \textit{negative} correlations with grammar and dictionary coverage, indicating that traditional documentation does not track performance on modern comprehension tasks. POS tagging is inconclusive: Stanza accuracy is ceiling-bound across the treebanked languages it supports, compressing the performance range. These patterns confirm that the dimensions carry meaningful, task-specific signal, and they motivate the empirically learned task weights in Table~\ref{tab:weights-learned}.

\begin{table}[t]
\centering
\resizebox{0.8\columnwidth}{!}{%
\begin{tabular}{lccccc}
\toprule
\textbf{Predictor} & \textbf{MT} & \textbf{BR} & \textbf{NLI} & \textbf{QA} \\
\midrule
Composite (CV)      & 0.37 & 0.24 & 0.20 & 0.65 \\
Best single dim.    & \textbf{0.49} & \textbf{0.47} & \textbf{0.36} & \textbf{0.69} \\
Parallel data only  & 0.46 & 0.41 & 0.34 & 0.69 \\
Log population      & 0.41 & 0.35 & 0.09 & 0.27 \\
EGIDS      & 0.34 & 0.23 & 0.18 & 0.55 \\
Equal-weight sum    & 0.28 & 0.20 & 0.23 & 0.29 \\
\bottomrule
\end{tabular}
}
\caption{Spearman's rank correlation ($\rho$) with task performance for the cross-validated composite and baseline predictors. POS tagging is omitted as inconclusive. The strongest predictor per task is in bold.}
\label{tab:composite-vs-baselines}
\end{table}

\paragraph{The composite does not beat the task-relevant dimension.} A natural question is whether combining all six dimensions predicts performance better than the single most relevant one. Learning task weights by cross-validated non-negative least squares (Appendix~\ref{sec:validation}), we find it does not (Table~\ref{tab:composite-vs-baselines}): across every task with sufficient range, the composite does not exceed the best single dimension (for MT, $\rho=0.37$ against $\rho=0.49$ for corpus coverage alone), though it consistently outperforms an equal-weight sum and simple baselines such as speaker population and vitality. This is a property of the dimensions rather than a failure of measurement. A principal component analysis over the 4,559 active languages shows the index is genuinely multidimensional; the first component explains only 54.7\% of variance, and four components are needed for 95\%, so the six dimensions are not reducible to a single latent axis. The one clear redundancy is between corpus and model coverage ($\rho=0.91$), reflecting their shared origin in web-crawling and pretraining pipelines; the remaining dimensions are only weakly intercorrelated ($\rho$ between $0.2$ and $0.5$). Parallel data is within 0.06 of the best single dimension on every task. This partly reflects the parallel test sets used for evaluation, and partly that parallel data acts as a proxy for other resources. Of the 690 languages with $P_i > 0$, 74.6\% have corpus and 80.7\% model coverage, against 22.5\% and 29.0\% overall. However, the parallel dimension cannot distinguish the 3{,}869 languages with $P_i = 0$ that have at least one other resource.

We therefore read ABAQUS as an instrument for characterizing the resource \textit{landscape}, where the multidimensional profile and the documentation-to-digitization gap are the object of interest, rather than as a single-number predictor of task readiness, for which the corresponding dimension should be used directly.

\begin{figure}[t]
    \centering
    \includegraphics[width=\columnwidth]{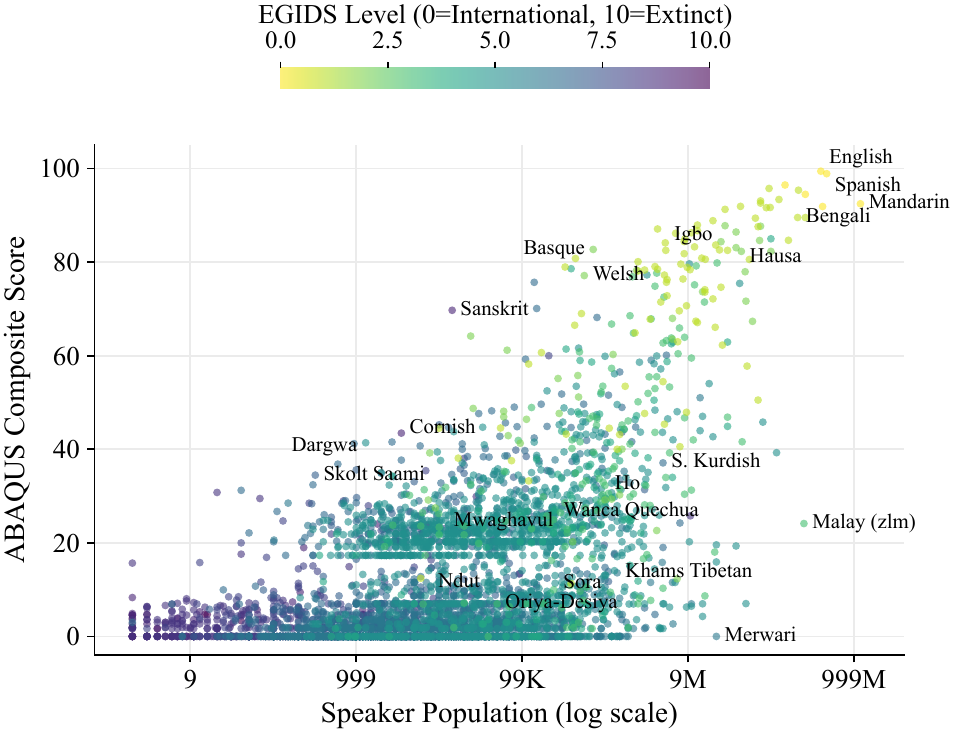}
    \caption{ABAQUS composite score plotted against speaker population (log scale). Marker color represents language vitality (EGIDS level), with lighter shades indicating more robust institutional support. A pseudo-count of +1 is applied to population to allow plotting of zero-speaker (extinct) languages.}
    \label{fig:population-disconnect}
\end{figure}

\subsection{Demographics and Vitality Status}
\label{sec:analysis-demographics}

To understand the relationship between a language's demographic profile and its technological standing, we compute the Spearman rank correlation between ABAQUS scores (default), speaker population, and institutional vitality defined by EGIDS\footnote{The Expanded Graded Intergenerational Disruption Scale (EGIDS) framework is a 13-level scale which scores languages based on vitality and endangerment. Scores range from 0 (International) to 10 (Extinct).} across the dataset, illustrated in Figure~\ref{fig:population-disconnect}. We find a moderate positive correlation between population and composite score ($\rho = 0.456$), and a moderate negative correlation between EGIDS level and composite score ($\rho = -0.445$, where lower EGIDS indicates higher vitality). 

While a larger speaker base generally correlates with a higher resource score, the variance is massive, proving that \textbf{population alone is a noisy and inadequate proxy for NLP resource level}. Additionally, we notice two distinct structural characteristics based on Figure~\ref{fig:population-disconnect}: Because grammars and dictionaries account for a maximum of 16.6\% of the composite score, languages that rely solely on traditional linguistic documentation cannot breach this threshold. To jump across the visible gap in the plot and score above 20, a language must cross the ``digital divide'' and be included in modern computational artifacts, such as a multilingual corpus or a language model. Additionally, outliers further underscore the disconnect between population and resources. While languages like Basque (\texttt{eus}) and Welsh (\texttt{cym}) score disproportionately high relative to their small populations due to dedicated institutional and community NLP efforts, other massive populations are starved of resources due to taxonomic artifacts. A striking example of this is Malay (\texttt{zlm}), which is attributed a population of roughly 250 million speakers, capturing the broader Malay macrolanguage continuum, yet achieves an ABAQUS score of only 24.12. This occurs because the NLP community overwhelmingly attributes data, models, and benchmarks from this continuum to the Indonesian (\texttt{ind}) ISO code, which ranks 11th overall, while \texttt{zlm} is starved of corpora and benchmark points. ``Low-resource'' status is thus not always a reflection of genuine data scarcity but in many cases is an artifact of taxonomic misalignment.

\section{Discussion and Conclusion}

While a substantial body of work has described the inequalities of language technology, these efforts have largely treated linguistic resources (such as grammars) and demographic factors (such as speaker population) in isolation, or foregrounded one at the expense of the other. What has been missing is a single, transparent index that situates a language's technological standing against both its resource profile and its demographic context. ABAQUS is a step toward filling this gap.

Applied to 8,168 languages across six dimensions, our index reveals a heavily right-skewed landscape in which nearly three-quarters of documented languages have no meaningful presence in contemporary NLP artifacts. Resource availability is neither uniform across dimensions nor reducible to a single axis. Grammars and dictionaries remain comparatively stable into the long tail, while benchmark coverage collapses almost entirely, exposing a documentation-to-digitization gap between languages that are nominally included in multilingual models and those whose capabilities are verifiably evaluated. Each downstream task in our validation process is predicted by a specific resource dimension, and while the composite is best read as a landscape-level summary rather than a per-task predictor, it consistently separates resource level from the coarse proxies of population and vitality alone.

We intend ABAQUS as an accessible and consultable reference: a platform on which speaker communities, developers, and policymakers can situate a given language within the current technological ecosystem. Crucially, ABAQUS is a descriptive instrument, not a prescriptive one. It issues no recommendations and calls for no particular action; as a resource index, it offers a clear and reproducible image of where a language stands, leaving the interpretation of that image, and any decisions that follow from it, to the communities and stakeholders best positioned to make them. 

\newpage
\section*{Limitations}

The sources underlying each dimension are manually curated and not exhaustive. The multilingual corpora, models, and benchmarks we include reflect what is publicly documented and disclosed at the time of collection, and are selected for being \textit{generally} multilingual rather than language-family specific. Resources that are not publicly cataloged, or that target a single family, are therefore not counted, and the index will require periodic updating as new resources appear. Matching also relies on ISO~639-3 codes as reported by each source; sources that label varieties inconsistently, or only by macrolanguage, may misattribute
coverage between closely related languages.

The index measures presence but not quality. A language listed among a model's supported languages, or present in a multilingual corpus or benchmark, is credited regardless of how much data underlies that presence, how well the resulting system performs, or whether the data is the result of automatic translation. As \citet{kreutzer-etal-2022-quality} show, lower-resource languages disproportionately suffer from low-quality data, so presence-based counts may overstate their true standing. ABAQUS should thus be read as a measure of how well a language \textit{could} be served by NLP, not how well it \textit{is} served.

Our validation is itself unevenly distributed across the resource spectrum. Because downstream evaluation depends on the existence of test data, the tasks we validate against are concentrated in the better-resourced part of the landscape: reading comprehension covers 112 languages, POS tagging only 71, and even the broadest tasks (MT, BR, and NLI) draw on Bible-derived data whose coverage, while wide, is genre-restricted. POS tagging is further limited by a ceiling effect across the treebanked languages Stanza supports, and we treat its correlations as inconclusive. Our findings on which dimensions predict which tasks should therefore be understood as strongest for the mid- to high-resource range, where evaluation data exists, and weakest precisely where the resource question is most acute.

Several analyses necessarily rest on subsamples rather than the full set of 8,168 languages. The principal component analysis and inter-dimension correlations are computed over the 4,559 \textit{active} (non-zero) languages, since the 3,609 all-zero languages carry no dimensional variation and would trivially inflate any measure of shared structure. Likewise, the demographic analysis is limited to languages for which population and vitality figures are available; these are missing or uncertain for many of the smallest languages, and the reported correlations should be read with that coverage gap in mind.

Finally, reducing six dimensions to a single composite necessarily obscures per-dimension detail: two languages with very different resource profiles can receive similar composite scores. This is a deliberate trade-off for a comparable, orderable quantity across all languages, and the underlying per-dimension scores remain available for finer-grained inspection. As our validation shows (\S\ref{sec:validation-body}), for anticipating performance on a particular task the relevant single dimension is preferable to the composite; the composite is best reserved for landscape-level comparison. For applications where preserving dimensional detail matters more than ranking, a non-aggregated notation such as that of \citet{coleman-etal-2026-ran} may be preferable.

\section*{Acknowledgments}
We would like to thank Pius von Däniken and Rico Sennrich for their constructive feedback on the design of ABAQUS.

\bibliography{custom}

@inproceedings{caswell-etal-2025-smol,
    title = "{SMOL}: Professionally Translated Parallel Data for 115 Under-represented Languages",
    author = "Caswell, Isaac  and
      Nielsen, Elizabeth  and
      Luo, Jiaming  and
      Cherry, Colin  and
      Kovacs, Geza  and
      Shemtov, Hadar  and
      Talukdar, Partha  and
      Tewari, Dinesh  and
      Diane, Baba Mamadi  and
      Diane, Djibrila  and
      Ciss{\'e}, Solo Farabado  and
      Doumbouya, Koulako Moussa  and
      Ferrante, Edoardo  and
      Guasoni, Alessandro  and
      Homan, Christopher  and
      Keita, Mamadou K.  and
      DebBarma, Sudhamoy  and
      Kuzhuget, Ali  and
      Anugraha, David  and
      Shulthan Habibi, Muhammad Ravi  and
      Ahmadi, Sina  and
      Munthali, Anthony  and
      Liu, Jonathan Mingfei  and
      Eng, Jonathan",
    editor = "Haddow, Barry  and
      Kocmi, Tom  and
      Koehn, Philipp  and
      Monz, Christof",
    booktitle = "Proceedings of the Tenth Conference on Machine Translation",
    month = nov,
    year = "2025",
    address = "Suzhou, China",
    publisher = "Association for Computational Linguistics",
    url = "https://aclanthology.org/2025.wmt-1.85/",
    doi = "10.18653/v1/2025.wmt-1.85",
    pages = "1103--1123",
    ISBN = "979-8-89176-341-8"
}

@inproceedings{rehm-etal-2020-european-language,
    title = "{E}uropean Language Grid: An Overview",
    author = "Rehm, Georg  and
      Berger, Maria  and
      Elsholz, Ela  and
      Hegele, Stefanie  and
      Kintzel, Florian  and
      Marheinecke, Katrin  and
      Piperidis, Stelios  and
      Deligiannis, Miltos  and
      Galanis, Dimitris  and
      Gkirtzou, Katerina  and
      Labropoulou, Penny  and
      Bontcheva, Kalina  and
      Jones, David  and
      Roberts, Ian  and
      Haji{\v{c}}, Jan  and
      Hamrlov{\'a}, Jana  and
      Ka{\v{c}}ena, Luk{\'a}{\v{s}}  and
      Choukri, Khalid  and
      Arranz, Victoria  and
      Vasi{\c{l}}jevs, Andrejs  and
      Anvari, Orians  and
      Lagzdi{\c{n}}{\v{s}}, Andis  and
      Me{\c{l}}{\c{n}}ika, J{\={u}}lija  and
      Backfried, Gerhard  and
      Dikici, Erin{\c{c}}  and
      Janosik, Miroslav  and
      Prinz, Katja  and
      Prinz, Christoph  and
      Stampler, Severin  and
      Thomas-Aniola, Dorothea  and
      G{\'o}mez-P{\'e}rez, Jos{\'e} Manuel  and
      Garcia Silva, Andres  and
      Berr{\'i}o, Christian  and
      Germann, Ulrich  and
      Renals, Steve  and
      Klejch, Ondrej",
    editor = "Calzolari, Nicoletta  and
      B{\'e}chet, Fr{\'e}d{\'e}ric  and
      Blache, Philippe  and
      Choukri, Khalid  and
      Cieri, Christopher  and
      Declerck, Thierry  and
      Goggi, Sara  and
      Isahara, Hitoshi  and
      Maegaard, Bente  and
      Mariani, Joseph  and
      Mazo, H{\'e}l{\`e}ne  and
      Moreno, Asuncion  and
      Odijk, Jan  and
      Piperidis, Stelios",
    booktitle = "Proceedings of the Twelfth Language Resources and Evaluation Conference",
    month = may,
    year = "2020",
    address = "Marseille, France",
    publisher = "European Language Resources Association",
    url = "https://aclanthology.org/2020.lrec-1.413/",
    pages = "3366--3380",
    language = "eng",
    ISBN = "979-10-95546-34-4"
}

@inproceedings{hedderich-etal-2021-survey,
    title = "A Survey on Recent Approaches for Natural Language Processing in Low-Resource Scenarios",
    author = {Hedderich, Michael A.  and
      Lange, Lukas  and
      Adel, Heike  and
      Str{\"o}tgen, Jannik  and
      Klakow, Dietrich},
    editor = "Toutanova, Kristina  and
      Rumshisky, Anna  and
      Zettlemoyer, Luke  and
      Hakkani-Tur, Dilek  and
      Beltagy, Iz  and
      Bethard, Steven  and
      Cotterell, Ryan  and
      Chakraborty, Tanmoy  and
      Zhou, Yichao",
    booktitle = "Proceedings of the 2021 Conference of the North American Chapter of the Association for Computational Linguistics: Human Language Technologies",
    month = jun,
    year = "2021",
    address = "Online",
    publisher = "Association for Computational Linguistics",
    url = "https://aclanthology.org/2021.naacl-main.201/",
    doi = "10.18653/v1/2021.naacl-main.201",
    pages = "2545--2568"
}

@inproceedings{bird-2020-decolonising,
    title = "Decolonising Speech and Language Technology",
    author = "Bird, Steven",
    editor = "Scott, Donia  and
      Bel, Nuria  and
      Zong, Chengqing",
    booktitle = "Proceedings of the 28th International Conference on Computational Linguistics",
    month = dec,
    year = "2020",
    address = "Barcelona, Spain (Online)",
    publisher = "International Committee on Computational Linguistics",
    url = "https://aclanthology.org/2020.coling-main.313/",
    doi = "10.18653/v1/2020.coling-main.313",
    pages = "3504--3519"
}

@inproceedings{coleman-etal-2026-ran,
    title = "{RAN}: Resource Abundance Notation for Languages in {NLP}",
    author = "Coleman, Jared  and
      Coleman, Tain{\~a}  and
      Krishnmachari, Bhaskar",
    editor = "Mager, Manuel  and
      Ebrahimi, Abteen  and
      Bui, Minh Duc  and
      Pugh, Robert  and
      Oncevay, Arturo  and
      Chiruzzo, Luis  and
      Solano, Rolando Coto  and
      Rijhwani, Shruti  and
      von der Wense, Katharina",
    booktitle = "Proceedings of the Sixth Workshop on {NLP} for Indigenous Languages of the {A}mericas ({A}mericas{NLP})",
    month = jul,
    year = "2026",
    address = "San Diego, California, USA",
    publisher = "Association for Computational Linguistics",
    url = "https://aclanthology.org/2026.americasnlp-6.15/",
    doi = "10.18653/v1/2026.americasnlp-6.15",
    pages = "168--172",
    ISBN = "979-8-89176-415-6"
}

@inproceedings{nordhoff2011glottolog,
  title={{Glottolog/Langdoc}: Defining dialects, languages, and language families as collections of resources},
  author={Nordhoff, Sebastian and Hammarstr{\"o}m, Harald},
  booktitle={First International Workshop on Linked Science 2011-In conjunction with the International Semantic Web Conference (ISWC 2011)},
  year={2011}
}

@inproceedings{zhang-etal-2024-hire,
    title = "Hire a Linguist!: Learning Endangered Languages in {LLM}s with In-Context Linguistic Descriptions",
    author = "Zhang, Kexun  and
      Choi, Yee Man  and
      Song, Zhenqiao  and
      He, Taiqi  and
      Wang, William Yang  and
      Li, Lei",
    editor = "Ku, Lun-Wei  and
      Martins, Andre  and
      Srikumar, Vivek",
    booktitle = "Findings of the Association for Computational Linguistics: ACL 2024",
    month = aug,
    year = "2024",
    address = "Bangkok, Thailand",
    publisher = "Association for Computational Linguistics",
    url = "https://aclanthology.org/2024.findings-acl.925/",
    doi = "10.18653/v1/2024.findings-acl.925",
    pages = "15654--15669"
}

@misc{platzgummer2026,
      title={\textit{Versteasch du mi?} Computational and Socio-Linguistic Perspectives on {GenAI, LLMs}, and Non-Standard Language}, 
      author={Verena Platzgummer and John McCrae and Sina Ahmadi},
      year={2026},
      eprint={2603.28213},
      archivePrefix={arXiv},
      primaryClass={cs.CL},
      url={https://arxiv.org/abs/2603.28213}, 
}

@inproceedings{gaspari-etal-2022-introducing,
    title = "Introducing the Digital Language Equality Metric: Technological Factors",
    author = "Gaspari, Federico  and
      Gallagher, Owen  and
      Rehm, Georg  and
      Giagkou, Maria  and
      Piperidis, Stelios  and
      Dunne, Jane  and
      Way, Andy",
    editor = "Aldabe, Itziar  and
      Altuna, Bego{\~n}a  and
      Farwell, Aritz  and
      Rigau, German",
    booktitle = "Proceedings of the Workshop Towards Digital Language Equality within the 13th Language Resources and Evaluation Conference",
    month = jun,
    year = "2022",
    address = "Marseille, France",
    publisher = "European Language Resources Association",
    url = "https://aclanthology.org/2022.tdle-1.1/",
    pages = "1--12"
}

@article{DBLP:journals/corr/abs-2103-09567,
  author       = {Mika H{\"{a}}m{\"{a}}l{\"{a}}inen},
  title        = {Endangered Languages are not Low-Resourced!},
  journal      = {CoRR},
  volume       = {abs/2103.09567},
  year         = {2021},
  url          = {https://arxiv.org/abs/2103.09567},
  eprinttype   = {arXiv},
  eprint       = {2103.09567},
  bibsource    = {dblp computer science bibliography, https://dblp.org}
}

@inproceedings{ranathunga-de-silva-2022-languages,
    title = "Some Languages are More Equal than Others: Probing Deeper into the Linguistic Disparity in the {NLP} World",
    author = "Ranathunga, Surangika  and
      de Silva, Nisansa",
    editor = "He, Yulan  and
      Ji, Heng  and
      Li, Sujian  and
      Liu, Yang  and
      Chang, Chua-Hui",
    booktitle = "Proceedings of the 2nd Conference of the Asia-Pacific Chapter of the Association for Computational Linguistics and the 12th International Joint Conference on Natural Language Processing (Volume 1: Long Papers)",
    month = nov,
    year = "2022",
    address = "Online only",
    publisher = "Association for Computational Linguistics",
    url = "https://aclanthology.org/2022.aacl-main.62/",
    doi = "10.18653/v1/2022.aacl-main.62",
    pages = "823--848"
}

@inproceedings{grutzner-zahn-rehm-2022-introducing,
    title = "Introducing the Digital Language Equality Metric: Contextual Factors",
    author = {Gr{\"u}tzner-Zahn, Annika  and
      Rehm, Georg},
    editor = "Aldabe, Itziar  and
      Altuna, Bego{\~n}a  and
      Farwell, Aritz  and
      Rigau, German",
    booktitle = "Proceedings of the Workshop Towards Digital Language Equality within the 13th Language Resources and Evaluation Conference",
    month = jun,
    year = "2022",
    address = "Marseille, France",
    publisher = "European Language Resources Association",
    url = "https://aclanthology.org/2022.tdle-1.2/",
    pages = "13--26"
}

@inproceedings{nigatu-etal-2024-zenos,
    title = "The {Z}eno{'}s Paradox of `Low-Resource' Languages",
    author = "Nigatu, Hellina Hailu  and
      Tonja, Atnafu Lambebo  and
      Rosman, Benjamin  and
      Solorio, Thamar  and
      Choudhury, Monojit",
    editor = "Al-Onaizan, Yaser  and
      Bansal, Mohit  and
      Chen, Yun-Nung",
    booktitle = "Proceedings of the 2024 Conference on Empirical Methods in Natural Language Processing",
    month = nov,
    year = "2024",
    address = "Miami, Florida, USA",
    publisher = "Association for Computational Linguistics",
    url = "https://aclanthology.org/2024.emnlp-main.983/",
    doi = "10.18653/v1/2024.emnlp-main.983",
    pages = "17753--17774"
}

@inproceedings{joshi-etal-2020-state,
    title = "The State and Fate of Linguistic Diversity and Inclusion in the {NLP} World",
    author = "Joshi, Pratik  and
      Santy, Sebastin  and
      Budhiraja, Amar  and
      Bali, Kalika  and
      Choudhury, Monojit",
    editor = "Jurafsky, Dan  and
      Chai, Joyce  and
      Schluter, Natalie  and
      Tetreault, Joel",
    booktitle = "Proceedings of the 58th Annual Meeting of the Association for Computational Linguistics",
    month = jul,
    year = "2020",
    address = "Online",
    publisher = "Association for Computational Linguistics",
    url = "https://aclanthology.org/2020.acl-main.560/",
    doi = "10.18653/v1/2020.acl-main.560",
    pages = "6282--6293"
}

@inproceedings{marone2026mmbert,
title={mm{BERT}: A Modern Multilingual Encoder with Annealed Language Learning},
author={Marc Marone and Orion Weller and William Fleshman and Eugene Yang and Dawn Lawrie and Benjamin Van Durme},
booktitle={Forty-third International Conference on Machine Learning},
year={2026},
url={https://openreview.net/forum?id=iJDJCO4mji}
}

@misc{swissai2025apertus,
  title={{Apertus: Democratizing Open and Compliant LLMs for Global Language Environments}},
  author={Alejandro Hernández-Cano and Alexander Hägele and Allen Hao Huang and Angelika Romanou and Antoni-Joan Solergibert and Barna Pasztor and Bettina Messmer and Dhia Garbaya and Eduard Frank Ďurech and Ido Hakimi and Juan García Giraldo and Mete Ismayilzada and Negar Foroutan and Skander Moalla and Tiancheng Chen and Vinko Sabolčec and Yixuan Xu and Michael Aerni and Badr AlKhamissi and Ines Altemir Marinas and Mohammad Hossein Amani and Matin Ansaripour and Ilia Badanin and Harold Benoit and Emanuela Boros and Nicholas Browning and Fabian Bösch and Maximilian Böther and Niklas Canova and Camille Challier and Clement Charmillot and Jonathan Coles and Jan Deriu and Arnout Devos and Lukas Drescher and Daniil Dzenhaliou and Maud Ehrmann and Dongyang Fan and Simin Fan and Silin Gao and Miguel Gila and María Grandury and Diba Hashemi and Alexander Hoyle and Jiaming Jiang and Mark Klein and Andrei Kucharavy and Anastasiia Kucherenko and Frederike Lübeck and Roman Machacek and Theofilos Manitaras and Andreas Marfurt and Kyle Matoba and Simon Matrenok and Henrique Mendoncça and Fawzi Roberto Mohamed and Syrielle Montariol and Luca Mouchel and Sven Najem-Meyer and Jingwei Ni and Gennaro Oliva and Matteo Pagliardini and Elia Palme and Andrei Panferov and Léo Paoletti and Marco Passerini and Ivan Pavlov and Auguste Poiroux and Kaustubh Ponkshe and Nathan Ranchin and Javi Rando and Mathieu Sauser and Jakhongir Saydaliev and Muhammad Ali Sayfiddinov and Marian Schneider and Stefano Schuppli and Marco Scialanga and Andrei Semenov and Kumar Shridhar and Raghav Singhal and Anna Sotnikova and Alexander Sternfeld and Ayush Kumar Tarun and Paul Teiletche and Jannis Vamvas and Xiaozhe Yao and Hao Zhao Alexander Ilic and Ana Klimovic and Andreas Krause and Caglar Gulcehre and David Rosenthal and Elliott Ash and Florian Tramèr and Joost VandeVondele and Livio Veraldi and Martin Rajman and Thomas Schulthess and Torsten Hoefler and Antoine Bosselut and Martin Jaggi and Imanol Schlag},
  year={2025},
  howpublished={\url{https://arxiv.org/abs/2509.14233}}
}

@inproceedings{ImaniGooghari_2023,
   title={Glot500: Scaling Multilingual Corpora and Language Models to 500 Languages},
   url={http://dx.doi.org/10.18653/v1/2023.acl-long.61},
   DOI={10.18653/v1/2023.acl-long.61},
   booktitle={Proceedings of the 61st Annual Meeting of the Association for Computational Linguistics (Volume 1: Long Papers)},
   publisher={Association for Computational Linguistics},
   author={ImaniGooghari, Ayyoob and Lin, Peiqin and Kargaran, Amir Hossein and Severini, Silvia and Jalili Sabet, Masoud and Kassner, Nora and Ma, Chunlan and Schmid, Helmut and Martins, André and Yvon, François and Schütze, Hinrich},
   year={2023},
   pages={1082–1117} }

@misc{omnilingualmtteam2026omnilingualmtmachinetranslation,
      title={Omnilingual {MT}: Machine Translation for 1,600 Languages}, 
      author={Omnilingual MT Team and Belen Alastruey and Niyati Bafna and Andrea Caciolai and Kevin Heffernan and Artyom Kozhevnikov and Christophe Ropers and Eduardo Sánchez and Charles-Eric Saint-James and Ioannis Tsiamas and Chierh Cheng and Joe Chuang and Paul-Ambroise Duquenne and Mark Duppenthaler and Nate Ekberg and Cynthia Gao and Pere Lluís Huguet Cabot and João Maria Janeiro and Jean Maillard and Gabriel Mejia Gonzalez and Holger Schwenk and Edan Toledo and Arina Turkatenko and Albert Ventayol-Boada and Rashel Moritz and Alexandre Mourachko and Surya Parimi and Mary Williamson and Shireen Yates and David Dale and Marta R. Costa-jussà},
      year={2026},
      eprint={2603.16309},
      archivePrefix={arXiv},
      primaryClass={cs.CL},
      url={https://arxiv.org/abs/2603.16309}, 
}

@article{DBLP:journals/nature/Team24,
  author       = {{NLLB Team}},
  title        = {Scaling neural machine translation to 200 languages},
  journal      = {Nat.},
  volume       = {630},
  number       = {8018},
  pages        = {841--846},
  year         = {2024},
  url          = {https://doi.org/10.1038/s41586-024-07335-x},
  doi          = {10.1038/S41586-024-07335-X},
  bibsource    = {dblp computer science bibliography, https://dblp.org}
}

@article{3648699.3648939,
    author = {Chowdhery, Aakanksha and Narang, Sharan and Devlin, Jacob and Bosma, Maarten and Mishra, Gaurav and Roberts, Adam and Barham, Paul and Chung, Hyung Won and Sutton, Charles and Gehrmann, Sebastian and Schuh, Parker and Shi, Kensen and Tsvyashchenko, Sashank and Maynez, Joshua and Rao, Abhishek and Barnes, Parker and Tay, Yi and Shazeer, Noam and Prabhakaran, Vinodkumar and Reif, Emily and Du, Nan and Hutchinson, Ben and Pope, Reiner and Bradbury, James and Austin, Jacob and Isard, Michael and Gur-Ari, Guy and Yin, Pengcheng and Duke, Toju and Levskaya, Anselm and Ghemawat, Sanjay and Dev, Sunipa and Michalewski, Henryk and Garcia, Xavier and Misra, Vedant and Robinson, Kevin and Fedus, Liam and Zhou, Denny and Ippolito, Daphne and Luan, David and Lim, Hyeontaek and Zoph, Barret and Spiridonov, Alexander and Sepassi, Ryan and Dohan, David and Agrawal, Shivani and Omernick, Mark and Dai, Andrew M. and Pillai, Thanumalayan Sankaranarayana and Pellat, Marie and Lewkowycz, Aitor and Moreira, Erica and Child, Rewon and Polozov, Oleksandr and Lee, Katherine and Zhou, Zongwei and Wang, Xuezhi and Saeta, Brennan and Diaz, Mark and Firat, Orhan and Catasta, Michele and Wei, Jason and Meier-Hellstern, Kathy and Eck, Douglas and Dean, Jeff and Petrov, Slav and Fiedel, Noah},
    title = {PaLM: scaling language modeling with pathways},
    year = {2023},
    issue_date = {January 2023},
    publisher = {JMLR.org},
    volume = {24},
    number = {1},
    issn = {1532-4435},
    journal = {J. Mach. Learn. Res.},
    month = jan,
    articleno = {240},
    numpages = {113}
}

@inproceedings{lin-etal-2022-shot,
    title = "Few-shot Learning with Multilingual Generative Language Models",
    author = "Lin, Xi Victoria  and
      Mihaylov, Todor  and
      Artetxe, Mikel  and
      Wang, Tianlu  and
      Chen, Shuohui  and
      Simig, Daniel  and
      Ott, Myle  and
      Goyal, Naman  and
      Bhosale, Shruti  and
      Du, Jingfei  and
      Pasunuru, Ramakanth  and
      Shleifer, Sam  and
      Koura, Punit Singh  and
      Chaudhary, Vishrav  and
      O{'}Horo, Brian  and
      Wang, Jeff  and
      Zettlemoyer, Luke  and
      Kozareva, Zornitsa  and
      Diab, Mona  and
      Stoyanov, Veselin  and
      Li, Xian",
    editor = "Goldberg, Yoav  and
      Kozareva, Zornitsa  and
      Zhang, Yue",
    booktitle = "Proceedings of the 2022 Conference on Empirical Methods in Natural Language Processing",
    month = dec,
    year = "2022",
    address = "Abu Dhabi, United Arab Emirates",
    publisher = "Association for Computational Linguistics",
    url = "https://aclanthology.org/2022.emnlp-main.616/",
    doi = "10.18653/v1/2022.emnlp-main.616",
    pages = "9019--9052"
}

@inproceedings{xue-etal-2021-mt5,
    title = "m{T}5: A Massively Multilingual Pre-trained Text-to-Text Transformer",
    author = "Xue, Linting  and
      Constant, Noah  and
      Roberts, Adam  and
      Kale, Mihir  and
      Al-Rfou, Rami  and
      Siddhant, Aditya  and
      Barua, Aditya  and
      Raffel, Colin",
    editor = "Toutanova, Kristina  and
      Rumshisky, Anna  and
      Zettlemoyer, Luke  and
      Hakkani-Tur, Dilek  and
      Beltagy, Iz  and
      Bethard, Steven  and
      Cotterell, Ryan  and
      Chakraborty, Tanmoy  and
      Zhou, Yichao",
    booktitle = "Proceedings of the 2021 Conference of the North American Chapter of the Association for Computational Linguistics: Human Language Technologies",
    month = jun,
    year = "2021",
    address = "Online",
    publisher = "Association for Computational Linguistics",
    url = "https://aclanthology.org/2021.naacl-main.41/",
    doi = "10.18653/v1/2021.naacl-main.41",
    pages = "483--498"
}

@inproceedings{conneau-etal-2020-unsupervised,
    title = "Unsupervised Cross-lingual Representation Learning at Scale",
    author = "Conneau, Alexis  and
      Khandelwal, Kartikay  and
      Goyal, Naman  and
      Chaudhary, Vishrav  and
      Wenzek, Guillaume  and
      Guzm{\'a}n, Francisco  and
      Grave, Edouard  and
      Ott, Myle  and
      Zettlemoyer, Luke  and
      Stoyanov, Veselin",
    editor = "Jurafsky, Dan  and
      Chai, Joyce  and
      Schluter, Natalie  and
      Tetreault, Joel",
    booktitle = "Proceedings of the 58th Annual Meeting of the Association for Computational Linguistics",
    month = jul,
    year = "2020",
    address = "Online",
    publisher = "Association for Computational Linguistics",
    url = "https://aclanthology.org/2020.acl-main.747/",
    doi = "10.18653/v1/2020.acl-main.747",
    pages = "8440--8451"
}

@article{JMLR:v25:23-0581,
  author  = {BigScience Workshop},
  title   = {{BLOOM}: A 176{B}-Parameter Open-Access Multilingual Language Model},
  journal = {Journal of Machine Learning Research},
  year    = {2024},
  volume  = {25},
  number  = {422},
  pages   = {1--74},
  url     = {http://jmlr.org/papers/v25/23-0581.html}
}

@misc{wei2023polylmopensourcepolyglot,
      title={{PolyLM}: An Open Source Polyglot Large Language Model}, 
      author={Xiangpeng Wei and Haoran Wei and Huan Lin and Tianhao Li and Pei Zhang and Xingzhang Ren and Mei Li and Yu Wan and Zhiwei Cao and Binbin Xie and Tianxiang Hu and Shangjie Li and Binyuan Hui and Bowen Yu and Dayiheng Liu and Baosong Yang and Fei Huang and Jun Xie},
      year={2023},
      eprint={2307.06018},
      archivePrefix={arXiv},
      primaryClass={cs.CL},
      url={https://arxiv.org/abs/2307.06018}, 
}

@misc{oepen2025hplt30largescalemultilingual,
      title={HPLT 3.0: Very Large-Scale Multilingual Resources for {LLM and MT. Mono- and Bi-lingual} Data, Multilingual Evaluation, and Pre-Trained Models}, 
      author={Stephan Oepen and Nikolay Arefev and Mikko Aulamo and Marta Bañón and Maja Buljan and Laurie Burchell and Lucas Charpentier and Pinzhen Chen and Mariya Fedorova and Ona de Gibert and Barry Haddow and Jan Hajič and Jindřich Helcl and Andrey Kutuzov and Veronika Laippala and Zihao Li and Risto Luukkonen and Bhavitvya Malik and Vladislav Mikhailov and Amanda Myntti and Dayyán O'Brien and Lucie Poláková and Sampo Pyysalo and Gema Ramírez Sánchez and Janine Siewert and Pavel Stepachev and Jörg Tiedemann and Teemu Vahtola and Dušan Variš and Fedor Vitiugin and Tea Vojtěchová and Jaume Zaragoza},
      year={2025},
      eprint={2511.01066},
      archivePrefix={arXiv},
      primaryClass={cs.CL},
      url={https://arxiv.org/abs/2511.01066}, 
}

@inproceedings{lauren_on2022the,
    title={The {BigScience} {ROOTS} Corpus: A 1.6{TB} Composite Multilingual Dataset},
    author={Hugo Lauren{\c{c}}on and Lucile Saulnier and Thomas Wang and Christopher Akiki and Albert Villanova del Moral and Teven Le Scao and Leandro Von Werra and Chenghao Mou and Eduardo Gonz{\'a}lez Ponferrada and Huu Nguyen and J{\"o}rg Frohberg and Mario {\v{S}}a{\v{s}}ko and Quentin Lhoest and Angelina McMillan-Major and G{\'e}rard Dupont and Stella Biderman and Anna Rogers and Loubna Ben allal and Francesco De Toni and Giada Pistilli and Olivier Nguyen and Somaieh Nikpoor and Maraim Masoud and Pierre Colombo and Javier de la Rosa and Paulo Villegas and Tristan Thrush and Shayne Longpre and Sebastian Nagel and Leon Weber and Manuel Romero Mu{\~n}oz and Jian Zhu and Daniel Van Strien and Zaid Alyafeai and Khalid Almubarak and Vu Minh Chien and Itziar Gonzalez-Dios and Aitor Soroa and Kyle Lo and Manan Dey and Pedro Ortiz Suarez and Aaron Gokaslan and Shamik Bose and David Ifeoluwa Adelani and Long Phan and Hieu Tran and Ian Yu and Suhas Pai and Jenny Chim and Violette Lepercq and Suzana Ilic and Margaret Mitchell and Sasha Luccioni and Yacine Jernite},
    booktitle={Thirty-sixth Conference on Neural Information Processing Systems Datasets and Benchmarks Track},
    year={2022},
    url={https://openreview.net/forum?id=UoEw6KigkUn}
}

@inproceedings{wenzek-etal-2020-cc100,
    title = "{CCN}et: Extracting High Quality Monolingual Datasets from Web Crawl Data",
    author = "Wenzek, Guillaume  and
      Lachaux, Marie-Anne  and
      Conneau, Alexis  and
      Chaudhary, Vishrav  and
      Guzm{\'a}n, Francisco  and
      Joulin, Armand  and
      Grave, Edouard",
    editor = "Calzolari, Nicoletta  and
      B{\'e}chet, Fr{\'e}d{\'e}ric  and
      Blache, Philippe  and
      Choukri, Khalid  and
      Cieri, Christopher  and
      Declerck, Thierry  and
      Goggi, Sara  and
      Isahara, Hitoshi  and
      Maegaard, Bente  and
      Mariani, Joseph  and
      Mazo, H{\'e}l{\`e}ne  and
      Moreno, Asuncion  and
      Odijk, Jan  and
      Piperidis, Stelios",
    booktitle = "Proceedings of the Twelfth Language Resources and Evaluation Conference",
    month = may,
    year = "2020",
    address = "Marseille, France",
    publisher = "European Language Resources Association",
    url = "https://aclanthology.org/2020.lrec-1.494/",
    pages = "4003--4012",
    language = "eng",
    ISBN = "979-10-95546-34-4"
}

@inproceedings{penedo2025fineweb,
    title={FineWeb2: One Pipeline to Scale Them All {\textemdash} Adapting Pre-Training Data Processing to Every Language},
    author={Guilherme Penedo and Hynek Kydl{\'\i}{\v{c}}ek and Vinko Sabol{\v{c}}ec and Bettina Messmer and Negar Foroutan and Amir Hossein Kargaran and Colin Raffel and Martin Jaggi and Leandro Von Werra and Thomas Wolf},
    booktitle={Second Conference on Language Modeling},
    year={2025},
    url={https://openreview.net/forum?id=jnRBe6zatP}
}

@article{raffel2023c4,
    author = {Raffel, Colin and Shazeer, Noam and Roberts, Adam and Lee, Katherine and Narang, Sharan and Matena, Michael and Zhou, Yanqi and Li, Wei and Liu, Peter J.},
    title = {Exploring the limits of transfer learning with a unified text-to-text transformer},
    year = {2020},
    issue_date = {January 2020},
    publisher = {JMLR.org},
    volume = {21},
    number = {1},
    issn = {1532-4435},
    journal = {J. Mach. Learn. Res.},
    month = jan,
    articleno = {140},
    numpages = {67},
    url = {https://dl.acm.org/doi/10.5555/3455716.3455856}
}

@inproceedings{kudugunta2023madlad,
    title={{MADLAD}-400: A Multilingual And Document-Level Large Audited Dataset},
    author={Sneha Kudugunta and Isaac Rayburn Caswell and Biao Zhang and Xavier Garcia and Derrick Xin and Aditya Kusupati and Romi Stella and Ankur Bapna and Orhan Firat},
    booktitle={Thirty-seventh Conference on Neural Information Processing Systems Datasets and Benchmarks Track},
    year={2023},
    url={https://openreview.net/forum?id=Y45ZCxslFx}
}

@inproceedings{schwenk-etal-2021-wikimatrix,
    title = "{W}iki{M}atrix: Mining 135{M} Parallel Sentences in 1620 Language Pairs from {W}ikipedia",
    author = "Schwenk, Holger  and
      Chaudhary, Vishrav  and
      Sun, Shuo  and
      Gong, Hongyu  and
      Guzm{\'a}n, Francisco",
    editor = "Merlo, Paola  and
      Tiedemann, Jorg  and
      Tsarfaty, Reut",
    booktitle = "Proceedings of the 16th Conference of the European Chapter of the Association for Computational Linguistics: Main Volume",
    month = apr,
    year = "2021",
    address = "Online",
    publisher = "Association for Computational Linguistics",
    url = "https://aclanthology.org/2021.eacl-main.115/",
    doi = "10.18653/v1/2021.eacl-main.115",
    pages = "1351--1361",
}

@inproceedings{lison-tiedemann-2016-opensubtitles2016,
    title = "{O}pen{S}ubtitles2016: Extracting Large Parallel Corpora from Movie and {TV} Subtitles",
    author = {Lison, Pierre  and
      Tiedemann, J{\"o}rg},
    editor = "Calzolari, Nicoletta  and
      Choukri, Khalid  and
      Declerck, Thierry  and
      Goggi, Sara  and
      Grobelnik, Marko  and
      Maegaard, Bente  and
      Mariani, Joseph  and
      Mazo, Helene  and
      Moreno, Asuncion  and
      Odijk, Jan  and
      Piperidis, Stelios",
    booktitle = "Proceedings of the Tenth International Conference on Language Resources and Evaluation ({LREC}'16)",
    month = may,
    year = "2016",
    address = "Portoro{\v{z}}, Slovenia",
    publisher = "European Language Resources Association (ELRA)",
    url = "https://aclanthology.org/L16-1147/",
    pages = "923--929"
}

@inproceedings{huang-etal-2025-benchmax,
    title = "{B}ench{MAX}: A Comprehensive Multilingual Evaluation Suite for Large Language Models",
    author = "Huang, Xu  and
      Zhu, Wenhao  and
      Hu, Hanxu  and
      He, Conghui  and
      Li, Lei  and
      Huang, Shujian  and
      Yuan, Fei",
    editor = "Christodoulopoulos, Christos  and
      Chakraborty, Tanmoy  and
      Rose, Carolyn  and
      Peng, Violet",
    booktitle = "Findings of the Association for Computational Linguistics: EMNLP 2025",
    month = nov,
    year = "2025",
    address = "Suzhou, China",
    publisher = "Association for Computational Linguistics",
    url = "https://aclanthology.org/2025.findings-emnlp.909/",
    doi = "10.18653/v1/2025.findings-emnlp.909",
    pages = "16751--16774",
    ISBN = "979-8-89176-335-7"
}

@inproceedings{xuan-etal-2025-mmlu,
    title = "{MMLU}-{P}ro{X}: A Multilingual Benchmark for Advanced Large Language Model Evaluation",
    author = "Xuan, Weihao  and
      Yang, Rui  and
      Qi, Heli  and
      Zeng, Qingcheng  and
      Xiao, Yunze  and
      Feng, Aosong  and
      Liu, Dairui  and
      Xing, Yun  and
      Wang, Junjue  and
      Gao, Fan  and
      Lu, Jinghui  and
      Jiang, Yuang  and
      Li, Huitao  and
      Li, Xin  and
      Yu, Kunyu  and
      Dong, Ruihai  and
      Gu, Shangding  and
      Li, Yuekang  and
      Xie, Xiaofei  and
      Juefei-Xu, Felix  and
      Khomh, Foutse  and
      Yoshie, Osamu  and
      Chen, Qingyu  and
      Teodoro, Douglas  and
      Liu, Nan  and
      Goebel, Randy  and
      Ma, Lei  and
      Marrese-Taylor, Edison  and
      Lu, Shijian  and
      Iwasawa, Yusuke  and
      Matsuo, Yutaka  and
      Li, Irene",
    editor = "Christodoulopoulos, Christos  and
      Chakraborty, Tanmoy  and
      Rose, Carolyn  and
      Peng, Violet",
    booktitle = "Proceedings of the 2025 Conference on Empirical Methods in Natural Language Processing",
    month = nov,
    year = "2025",
    address = "Suzhou, China",
    publisher = "Association for Computational Linguistics",
    url = "https://aclanthology.org/2025.emnlp-main.79/",
    doi = "10.18653/v1/2025.emnlp-main.79",
    pages = "1513--1532",
    ISBN = "979-8-89176-332-6"
}

@inproceedings{hofman-etal-2026-maps,
    title = "{MAPS}: A Multilingual Benchmark for Agent Performance and Security",
    author = "Hofman, Omer  and
      Brokman, Jonathan  and
      Rachmil, Oren  and
      Bose, Shamik  and
      Pahuja, Vikas  and
      Shimizu, Toshiya  and
      Starostina, Trisha  and
      Marchisio, Kelly  and
      Goldfarb-Tarrant, Seraphina  and
      Vainshtein, Roman",
    editor = "Demberg, Vera  and
      Inui, Kentaro  and
      Marquez, Llu{\'i}s",
    booktitle = "Findings of the {A}ssociation for {C}omputational {L}inguistics: {EACL} 2026",
    month = mar,
    year = "2026",
    address = "Rabat, Morocco",
    publisher = "Association for Computational Linguistics",
    url = "https://aclanthology.org/2026.findings-eacl.42/",
    doi = "10.18653/v1/2026.findings-eacl.42",
    pages = "821--845",
    ISBN = "979-8-89176-386-9"
}

@inproceedings{hu2020xtrememassivelymultilingualmultitask,
    author = {Hu, Junjie and Ruder, Sebastian and Siddhant, Aditya and Neubig, Graham and Firat, Orhan and Johnson, Melvin},
    title = {{XTREME}: a massively multilingual multi-task benchmark for evaluating cross-lingual generalization},
    year = {2020},
    publisher = {JMLR.org},
    booktitle = {Proceedings of the 37th International Conference on Machine Learning},
    articleno = {410},
    numpages = {11},
    series = {ICML'20},
    url = {https://dl.acm.org/doi/10.5555/3524938.3525348}
}

@article{gayol2021flores,
    title = "The {F}lores-101 Evaluation Benchmark for Low-Resource and Multilingual Machine Translation",
    author = "Goyal, Naman  and
      Gao, Cynthia  and
      Chaudhary, Vishrav  and
      Chen, Peng-Jen  and
      Wenzek, Guillaume  and
      Ju, Da  and
      Krishnan, Sanjana  and
      Ranzato, Marc{'}Aurelio  and
      Guzm{\'a}n, Francisco  and
      Fan, Angela",
    editor = "Roark, Brian  and
      Nenkova, Ani",
    journal = "Transactions of the Association for Computational Linguistics",
    volume = "10",
    year = "2022",
    address = "Cambridge, MA",
    publisher = "MIT Press",
    url = "https://aclanthology.org/2022.tacl-1.30/",
    doi = "10.1162/tacl_a_00474",
    pages = "522--538"
}

@misc{vayani2025alm,
      title={All Languages Matter: Evaluating {LMMs} on Culturally Diverse 100 Languages}, 
      author={Ashmal Vayani and Dinura Dissanayake and Hasindri Watawana and Noor Ahsan and Nevasini Sasikumar and Omkar Thawakar and Henok Biadglign Ademtew and Yahya Hmaiti and Amandeep Kumar and Kartik Kuckreja and Mykola Maslych and Wafa Al Ghallabi and Mihail Mihaylov and Chao Qin and Abdelrahman M Shaker and Mike Zhang and Mahardika Krisna Ihsani and Amiel Esplana and Monil Gokani and Shachar Mirkin and Harsh Singh and Ashay Srivastava and Endre Hamerlik and Fathinah Asma Izzati and Fadillah Adamsyah Maani and Sebastian Cavada and Jenny Chim and Rohit Gupta and Sanjay Manjunath and Kamila Zhumakhanova and Feno Heriniaina Rabevohitra and Azril Amirudin and Muhammad Ridzuan and Daniya Kareem and Ketan More and Kunyang Li and Pramesh Shakya and Muhammad Saad and Amirpouya Ghasemaghaei and Amirbek Djanibekov and Dilshod Azizov and Branislava Jankovic and Naman Bhatia and Alvaro Cabrera and Johan Obando-Ceron and Olympiah Otieno and Fabian Farestam and Muztoba Rabbani and Sanoojan Baliah and Santosh Sanjeev and Abduragim Shtanchaev and Maheen Fatima and Thao Nguyen and Amrin Kareem and Toluwani Aremu and Nathan Xavier and Amit Bhatkal and Hawau Toyin and Aman Chadha and Hisham Cholakkal and Rao Muhammad Anwer and Michael Felsberg and Jorma Laaksonen and Thamar Solorio and Monojit Choudhury and Ivan Laptev and Mubarak Shah and Salman Khan and Fahad Khan},
      year={2025},
      eprint={2411.16508},
      archivePrefix={arXiv},
      primaryClass={cs.CV},
      url={https://arxiv.org/abs/2411.16508}, 
}

@inproceedings{zevallos-etal-2024-tema,
    title = "{TEMA}: Token Embeddings Mapping for Enriching Low-Resource Language Models",
    author = "Zevallos, Rodolfo  and
      Bel, N{\'u}ria  and
      Farr{\'u}s, Mireia",
    editor = "Al-Onaizan, Yaser  and
      Bansal, Mohit  and
      Chen, Yun-Nung",
    booktitle = "Proceedings of the 2024 Conference on Empirical Methods in Natural Language Processing",
    month = nov,
    year = "2024",
    address = "Miami, Florida, USA",
    publisher = "Association for Computational Linguistics",
    url = "https://aclanthology.org/2024.emnlp-main.638/",
    doi = "10.18653/v1/2024.emnlp-main.638",
    pages = "11423--11435"
}

@inproceedings{singh-2008-natural,
    title = "Natural Language Processing for Less Privileged Languages: Where do we come from? Where are we going?",
    author = "Singh, Anil Kumar",
    booktitle = "Proceedings of the {IJCNLP}-08 Workshop on {NLP} for Less Privileged Languages",
    year = "2008",
    url = "https://aclanthology.org/I08-3004/"
}

@inproceedings{knowles-littell-2022-translation,
    title = "Translation Memories as Baselines for Low-Resource Machine Translation",
    author = "Knowles, Rebecca and Littell, Patrick",
    editor = "Calzolari, Nicoletta and B{\'e}chet, Fr{\'e}d{\'e}ric and Blache, Philippe and Choukri, Khalid and Cieri, Christopher  and
      Declerck, Thierry  and
      Goggi, Sara  and
      Isahara, Hitoshi  and
      Maegaard, Bente  and
      Mariani, Joseph  and
      Mazo, H{\'e}l{\`e}ne  and
      Odijk, Jan  and
      Piperidis, Stelios",
    booktitle = "Proceedings of the Thirteenth Language Resources and Evaluation Conference",
    month = jun,
    year = "2022",
    address = "Marseille, France",
    publisher = "European Language Resources Association",
    url = "https://aclanthology.org/2022.lrec-1.728/",
    pages = "6759--6767",
}

@inproceedings{sefat2026glotweb,
  title={Glot{W}eb: Web Indexing for Minority Languages},
  author={Sefat, Abdullah Al and Kargaran, Amir Hossein and Yvon, Fran{\c{c}}ois and Sch{\"u}tze, Hinrich},
  booktitle={Proceedings of the ACM Web Conference 2026},
  pages={8469--8472},
  year={2026},
  url={https://dl.acm.org/doi/abs/10.1145/3774904.3792887}
}

@book{moseley2010atlas,
  title        = {Atlas of the World's Languages in Danger},
  editor       = {Christopher Moseley},
  author       = {{UNESCO}},
  year         = {2010},
  publisher    = {UNESCO Publishing},
  address      = {Paris},
  isbn         = {978-92-3-104096-2},
  series       = {Memory of Peoples},
  language     = {English},
  note         = {Cartography by Alexandre Nicolas. Also available in French and Spanish. Licensed under CC BY-SA 3.0 IGO}
}

@inproceedings{mehta2020gondi,
  title     = {Learnings from Technological Interventions in a Low Resource Language: A Case-Study on {Gondi}},
  author    = {Mehta, Devansh and Santy, Sebastin and Mothilal, Ramaravind Kommiya and Srivastava, Brij Mohan Lal and Sharma, Alok and Shukla, Anurag and Prasad, Vishnu and Venkanna, U. and Sharma, Amit and Bali, Kalika},
  booktitle = {Proceedings of the Twelfth Language Resources and Evaluation Conference},
  pages     = {2832--2838},
  year      = {2020},
  address   = {Marseille, France},
  publisher = {European Language Resources Association},
  url       = {https://aclanthology.org/2020.lrec-1.345/}
}

@inproceedings{blasi-etal-2022-systematic,
    title = "Systematic Inequalities in Language Technology Performance across the World{'}s Languages",
    author = "Blasi, Damian  and
      Anastasopoulos, Antonios  and
      Neubig, Graham",
    editor = "Muresan, Smaranda  and
      Nakov, Preslav  and
      Villavicencio, Aline",
    booktitle = "Proceedings of the 60th Annual Meeting of the Association for Computational Linguistics (Volume 1: Long Papers)",
    month = may,
    year = "2022",
    address = "Dublin, Ireland",
    publisher = "Association for Computational Linguistics",
    url = "https://aclanthology.org/2022.acl-long.376/",
    doi = "10.18653/v1/2022.acl-long.376",
    pages = "5486--5505"
}

@inproceedings{popovic-2015-chrf,
    title = "chr{F}: character n-gram {F}-score for automatic {MT} evaluation",
    author = "Popovi{\'c}, Maja",
    editor = "Bojar, Ond{\v{r}}ej  and
      Chatterjee, Rajan  and
      Federmann, Christian  and
      Haddow, Barry  and
      Hokamp, Chris  and
      Huck, Matthias  and
      Logacheva, Varvara  and
      Pecina, Pavel",
    booktitle = "Proceedings of the Tenth Workshop on Statistical Machine Translation",
    month = sep,
    year = "2015",
    address = "Lisbon, Portugal",
    publisher = "Association for Computational Linguistics",
    url = "https://aclanthology.org/W15-3049/",
    doi = "10.18653/v1/W15-3049",
    pages = "392--395"
}

@article{kreutzer-etal-2022-quality,
    title = "Quality at a Glance: An Audit of Web-Crawled Multilingual Datasets",
    author = {Kreutzer, Julia  and
      Caswell, Isaac  and
      Wang, Lisa  and
      Wahab, Ahsan  and
      van Esch, Daan  and
      Ulzii-Orshikh, Nasanbayar  and
      Tapo, Allahsera  and
      Subramani, Nishant  and
      Sokolov, Artem  and
      Sikasote, Claytone  and
      Setyawan, Monang  and
      Sarin, Supheakmungkol  and
      Samb, Sokhar  and
      Sagot, Beno{\^i}t  and
      Rivera, Clara  and
      Rios, Annette  and
      Papadimitriou, Isabel  and
      Osei, Salomey  and
      Suarez, Pedro Ortiz  and
      Orife, Iroro  and
      Ogueji, Kelechi  and
      Rubungo, Andre Niyongabo  and
      Nguyen, Toan Q.  and
      M{\"u}ller, Mathias  and
      M{\"u}ller, Andr{\'e}  and
      Muhammad, Shamsuddeen Hassan  and
      Muhammad, Nanda  and
      Mnyakeni, Ayanda  and
      Mirzakhalov, Jamshidbek  and
      Matangira, Tapiwanashe  and
      Leong, Colin  and
      Lawson, Nze  and
      Kudugunta, Sneha  and
      Jernite, Yacine  and
      Jenny, Mathias  and
      Firat, Orhan  and
      Dossou, Bonaventure F. P.  and
      Dlamini, Sakhile  and
      de Silva, Nisansa  and
      {\c{C}}abuk Ball{\i}, Sakine  and
      Biderman, Stella  and
      Battisti, Alessia  and
      Baruwa, Ahmed  and
      Bapna, Ankur  and
      Baljekar, Pallavi  and
      Azime, Israel Abebe  and
      Awokoya, Ayodele  and
      Ataman, Duygu  and
      Ahia, Orevaoghene  and
      Ahia, Oghenefego  and
      Agrawal, Sweta  and
      Adeyemi, Mofetoluwa},
    editor = "Roark, Brian  and
      Nenkova, Ani",
    journal = "Transactions of the Association for Computational Linguistics",
    volume = "10",
    year = "2022",
    address = "Cambridge, MA",
    publisher = "MIT Press",
    url = "https://aclanthology.org/2022.tacl-1.4/",
    doi = "10.1162/tacl_a_00447",
    pages = "50--72"
}

@book{glottolog5.3,
  author = {Hammarström, Harald and Forkel, Robert and Haspelmath, Martin and Bank, Sebastian},
  title = {Glottolog 5.3},
  year = {2026},
  publisher = {Max Planck Institute for Evolutionary Anthropology},
  url = {http://glottolog.org}
}

@inproceedings{tiedemann-2012-parallel,
    title = "Parallel Data, Tools and Interfaces in {OPUS}",
    author = {Tiedemann, J{\"o}rg},
    editor = "Calzolari, Nicoletta  and
      Choukri, Khalid  and
      Declerck, Thierry  and
      Do{\u{g}}an, Mehmet U{\u{g}}ur  and
      Maegaard, Bente  and
      Mariani, Joseph  and
      Moreno, Asuncion  and
      Odijk, Jan  and
      Piperidis, Stelios",
    booktitle = "Proceedings of the Eighth International Conference on Language Resources and Evaluation ({LREC}'12)",
    month = may,
    year = "2012",
    address = "Istanbul, Turkey",
    publisher = "European Language Resources Association (ELRA)",
    url = "https://aclanthology.org/L12-1246/",
    pages = "2214--2218"
}

@inproceedings{DBLP:conf/chi/NigatuCC24,
  author       = {Hellina Hailu Nigatu and
                  John F. Canny and
                  Sarah E. Chasins},
  editor       = {Florian 'Floyd' Mueller and
                  Penny Kyburz and
                  Julie R. Williamson and
                  Corina Sas and
                  Max L. Wilson and
                  Phoebe O. Toups Dugas and
                  Irina Shklovski},
  title        = {Low-Resourced Languages and Online Knowledge Repositories: {A} Need-Finding
                  Study},
  booktitle    = {Proceedings of the {CHI} Conference on Human Factors in Computing
                  Systems, {CHI} 2024, Honolulu, HI, USA, May 11-16, 2024},
  pages        = {562:1--562:21},
  publisher    = {{ACM}},
  year         = {2024},
  url          = {https://doi.org/10.1145/3613904.3642605},
  doi          = {10.1145/3613904.3642605},
  bibsource    = {dblp computer science bibliography, https://dblp.org}
}

@inproceedings{bird-2026-big,
    title = "Big {AI} is Accelerating the Metacrisis: What Can We Do?",
    author = "Bird, Steven",
    editor = "Liakata, Maria  and
      Moreira, Viviane P.  and
      Zhang, Jiajun  and
      Jurgens, David",
    booktitle = "Proceedings of the 64th Annual Meeting of the {A}ssociation for {C}omputational {L}inguistics (Volume 2: Short Papers)",
    month = jul,
    year = "2026",
    address = "San Diego, California, United States",
    publisher = "Association for Computational Linguistics",
    url = "https://aclanthology.org/2026.acl-short.6/",
    doi = "10.18653/v1/2026.acl-short.6",
    pages = "50--61",
    ISBN = "979-8-89176-391-3"
}

@inproceedings{liu-etal-2022-always,
    title = "Not always about you: Prioritizing community needs when developing endangered language technology",
    author = "Liu, Zoey  and
      Richardson, Crystal  and
      Hatcher, Richard  and
      Prud{'}hommeaux, Emily",
    editor = "Muresan, Smaranda  and
      Nakov, Preslav  and
      Villavicencio, Aline",
    booktitle = "Proceedings of the 60th Annual Meeting of the Association for Computational Linguistics (Volume 1: Long Papers)",
    month = may,
    year = "2022",
    address = "Dublin, Ireland",
    publisher = "Association for Computational Linguistics",
    url = "https://aclanthology.org/2022.acl-long.272/",
    doi = "10.18653/v1/2022.acl-long.272",
    pages = "3933--3944"
}

@inproceedings{qaddoumi-etal-2026-learning,
    title = "Learning Stress in {A}rabic Low-Resource Settings",
    author = "Qaddoumi, Abed  and
      Kodner, Jordan  and
      Rambow, Owen  and
      Khalifa, Salam  and
      Heinz, Jeffrey",
    editor = "Voigt, Rob  and
      Warstadt, Alex  and
      Feldman, Naomi  and
      Linzen, Tal",
    booktitle = "Proceedings of the Society for Computation in Linguistics 2026",
    month = jul,
    year = "2026",
    address = "San Diego, CA",
    publisher = "Association for Computational Linguistics",
    url = "https://aclanthology.org/2026.scil-main.24/",
    doi = "10.18653/v1/2026.scil-main.24",
    pages = "262--279",
    ISBN = "979-8-89176-412-5"
}

@inproceedings{DBLP:conf/acl/Bird22,
  author       = {Steven Bird},
  editor       = {Smaranda Muresan and
                  Preslav Nakov and
                  Aline Villavicencio},
  title        = {Local Languages, Third Spaces, and other High-Resource Scenarios},
  booktitle    = {Proceedings of the 60th Annual Meeting of the Association for Computational
                  Linguistics (Volume 1: Long Papers), {ACL} 2022, Dublin, Ireland,
                  May 22-27, 2022},
  pages        = {7817--7829},
  publisher    = {Association for Computational Linguistics},
  year         = {2022},
  url          = {https://doi.org/10.18653/v1/2022.acl-long.539},
  doi          = {10.18653/V1/2022.ACL-LONG.539},
  bibsource    = {dblp computer science bibliography, https://dblp.org}
}

@inproceedings{DBLP:conf/emnlp/SainzPELAGZARAAS25,
  author       = {Oscar Sainz and
                  Naiara P{\'{e}}rez and
                  Julen Etxaniz and
                  Joseba Fernandez de Landa and
                  Itziar Aldabe and
                  Iker Garc{\'{\i}}a{-}Ferrero and
                  Aimar Zabala and
                  Ekhi Azurmendi and
                  German Rigau and
                  Eneko Agirre and
                  Mikel Artetxe and
                  Aitor Soroa},
  editor       = {Christos Christodoulopoulos and
                  Tanmoy Chakraborty and
                  Carolyn Rose and
                  Violet Peng},
  title        = {Instructing Large Language Models for Low-Resource Languages: {A}
                  Systematic Study for {B}asque},
  booktitle    = {Proceedings of the 2025 Conference on Empirical Methods in Natural
                  Language Processing, {EMNLP} 2025, Suzhou, China, November 4-9, 2025},
  pages        = {29136--29160},
  publisher    = {Association for Computational Linguistics},
  year         = {2025},
  url          = {https://doi.org/10.18653/v1/2025.emnlp-main.1484},
  doi          = {10.18653/V1/2025.EMNLP-MAIN.1484},
  bibsource    = {dblp computer science bibliography, https://dblp.org}
}

@inproceedings{gagnier-kirubakaran-2026-deer,
    title = "Deer, Deities, and Dancing: Culturally Biased {LLM} Hallucination in Low-Resource {W}ix{\'a}rika Translation",
    author = "Gagnier, Henry  and
      Kirubakaran, Ashwin",
    editor = "Mager, Manuel  and
      Ebrahimi, Abteen  and
      Bui, Minh Duc  and
      Pugh, Robert  and
      Oncevay, Arturo  and
      Chiruzzo, Luis  and
      Solano, Rolando Coto  and
      Rijhwani, Shruti  and
      von der Wense, Katharina",
    booktitle = "Proceedings of the Sixth Workshop on {NLP} for Indigenous Languages of the {A}mericas ({A}mericas{NLP})",
    month = jul,
    year = "2026",
    address = "San Diego, California, USA",
    publisher = "Association for Computational Linguistics",
    url = "https://aclanthology.org/2026.americasnlp-6.7/",
    doi = "10.18653/v1/2026.americasnlp-6.7",
    pages = "74--81",
    ISBN = "979-8-89176-415-6"
}

@inproceedings{ahmadi-etal-2025-parme,
    title = "{PARME}: Parallel Corpora for Low-Resourced {M}iddle {E}astern Languages",
    author = {Ahmadi, Sina  and
      Sennrich, Rico  and
      Karami, Erfan  and
      Marani, Ako  and
      Fekrazad, Parviz  and
      Baghban, Gholamreza Akbarzadeh  and
      Hadi, Hanah  and
      Heidari, Semko  and
      Dogan, Mah{\^i}r  and
      Asadi, Pedram  and
      Bashir, Dashne  and
      Ghodrati, Mohammad Amin  and
      Amini, Kourosh  and
      Ashourinezhad, Zeynab  and
      Baladi, Mana  and
      Ezzati, Farshid  and
      Ghasemifar, Alireza  and
      Hosseinpour, Daryoush  and
      Abbaszadeh, Behrooz  and
      Hassanpour, Amin  and
      Hamaamin, Bahaddin Jalal  and
      Hama, Saya Kamal  and
      Mousavi, Ardeshir  and
      Hussein, Sarko Nazir  and
      Nejadgholi, Isar  and
      {\"O}lmez, Mehmet  and
      Osmanpour, Horam  and
      Ramezani, Rashid Roshan  and
      Aziz, Aryan Sediq  and
      Salehi Sheikhalikelayeh, Ali  and
      Yadegari, Mohammadreza  and
      Yadegari, Kewyar  and
      Roodsari, Sedighe Zamani},
    editor = "Che, Wanxiang  and
      Nabende, Joyce  and
      Shutova, Ekaterina  and
      Pilehvar, Mohammad Taher",
    booktitle = "Proceedings of the 63rd Annual Meeting of the Association for Computational Linguistics (Volume 1: Long Papers)",
    month = jul,
    year = "2025",
    address = "Vienna, Austria",
    publisher = "Association for Computational Linguistics",
    url = "https://aclanthology.org/2025.acl-long.1451/",
    doi = "10.18653/v1/2025.acl-long.1451",
    pages = "30032--30053",
    ISBN = "979-8-89176-251-0"
}

@article{bender-friedman-2018-data,
    title = "Data Statements for Natural Language Processing: Toward Mitigating System Bias and Enabling Better Science",
    author = "Bender, Emily M.  and
      Friedman, Batya",
    editor = "Lee, Lillian  and
      Johnson, Mark  and
      Toutanova, Kristina  and
      Roark, Brian",
    journal = "Transactions of the Association for Computational Linguistics",
    volume = "6",
    year = "2018",
    address = "Cambridge, MA",
    publisher = "MIT Press",
    url = "https://aclanthology.org/Q18-1041/",
    doi = "10.1162/tacl_a_00041",
    pages = "587--604"
}

@inproceedings{cieri-etal-2016-selection,
    title = "Selection Criteria for Low Resource Language Programs",
    author = "Cieri, Christopher  and
      Maxwell, Mike  and
      Strassel, Stephanie  and
      Tracey, Jennifer",
    editor = "Calzolari, Nicoletta  and
      Choukri, Khalid  and
      Declerck, Thierry  and
      Goggi, Sara  and
      Grobelnik, Marko  and
      Maegaard, Bente  and
      Mariani, Joseph  and
      Mazo, Helene  and
      Moreno, Asuncion  and
      Odijk, Jan  and
      Piperidis, Stelios",
    booktitle = "Proceedings of the Tenth International Conference on Language Resources and Evaluation ({LREC}'16)",
    month = may,
    year = "2016",
    address = "Portoro{\v{z}}, Slovenia",
    publisher = "European Language Resources Association (ELRA)",
    url = "https://aclanthology.org/L16-1720/",
    pages = "4543--4549"
}

@article{lex2014upset,
  title={{UpSet}: visualization of intersecting sets},
  author={Lex, Alexander and Gehlenborg, Nils and Strobelt, Hendrik and Vuillemot, Romain and Pfister, Hanspeter},
  journal={IEEE transactions on visualization and computer graphics},
  volume={20},
  number={12},
  pages={1983--1992},
  year={2014},
  publisher={IEEE}
}

@inproceedings{post-2018-call,
    title = "A Call for Clarity in Reporting {BLEU} Scores",
    author = "Post, Matt",
    editor = "Bojar, Ond{\v{r}}ej  and
      Chatterjee, Rajen  and
      Federmann, Christian  and
      Fishel, Mark  and
      Graham, Yvette  and
      Haddow, Barry  and
      Huck, Matthias  and
      Yepes, Antonio Jimeno  and
      Koehn, Philipp  and
      Monz, Christof  and
      Negri, Matteo  and
      N{\'e}v{\'e}ol, Aur{\'e}lie  and
      Neves, Mariana  and
      Post, Matt  and
      Specia, Lucia  and
      Turchi, Marco  and
      Verspoor, Karin",
    booktitle = "Proceedings of the Third Conference on Machine Translation: Research Papers",
    month = oct,
    year = "2018",
    address = "Brussels, Belgium",
    publisher = "Association for Computational Linguistics",
    url = "https://aclanthology.org/W18-6319/",
    doi = "10.18653/v1/W18-6319",
    pages = "186--191"
}

@article{wang2024multilingual,
  title={Multilingual {E5} text embeddings: A technical report},
  author={Wang, Liang and Yang, Nan and Huang, Xiaolong and Yang, Linjun and Majumder, Rangan and Wei, Furu},
  journal={arXiv preprint arXiv:2402.05672},
  year={2024}
}

\clearpage
\appendix

\section{Validation of Resource Dimensions}
\label{sec:validation}

To assess whether the resource dimensions underlying ABAQUS reflect genuine technological disparity, and to derive the task-specific weight configurations reported in Table~\ref{tab:weights-learned}, we correlate the per-dimension scores against observed downstream performance across five tasks: MT, BR, NLI, reading comprehension, and POS tagging.

\paragraph{MT and BR.} We evaluate translation into and out of English on a held-out set of Bible verses, which affords parallel coverage across a far larger set of languages than Flores-200. For each language we report chrF~\citep{popovic-2015-chrf} averaged over both directions, computed with SacreBLEU~\citep{post-2018-call}, for three systems: NLLB-200 (3.3B)~\citep{DBLP:journals/nature/Team24}, MADLAD-400 (10B)~\citep{kudugunta2023madlad}, and Apertus (8B)~\citep{swissai2025apertus}. For BR, we report top-1 accuracy over the same parallel verses using seven multilingual sentence encoders, including LaBSE~\citep{DBLP:conf/acl/FengYCA022}, BGE-M3~\citep{DBLP:conf/acl/ChenXZLLL24}, and the multilingual E5~\citep{wang2024multilingual} and GTE~\citep{DBLP:conf/emnlp/ZhangZLXDTLYXHZ24} families. Both tasks cover 807 languages.

\paragraph{NLI and reading comprehension.} For NLI, we report accuracy over 833 languages using three XNLI\citep{conneau2018xnli}-finetuned models (mDeBERTa~\citep{DBLP:conf/iclr/HeGC23}, XLM-R large, and mT5) applied to Bible-derived entailment pairs. For reading comprehension we use the published per-language results of Belebele \citep{DBLP:conf/acl/BandarkarLMASHG24}, a parallel multiple-choice benchmark, covering 112 of our languages across eleven model and evaluation settings.

\paragraph{POS tagging.} We report UPOS accuracy on the test splits of Universal Dependencies~2.15~\citep{andrews-coltekin-2025-developing} using off-the-shelf Stanza pipelines \citep{qi-etal-2020-stanza}. Because Stanza provides pretrained models for a limited set of treebanked languages, this task covers 71 languages, concentrated in the higher-resource range.

\section{Sensitivity Analysis}
\label{app:sensitivity}

The paper's main results use one default weighting (Table~\ref{tab:weights}). These defaults represent a design choice rather than a learned scheme as in Table~\ref{tab:weights-learned}, which raises the question: do the main findings hold under alternative weighting? To answer this, we recompute the composite score over alternative weightings and measure how much the language scores and ranks change. Three types of alternatives are tested: (i)~fixed schemes, namely equal weights, documentation only removed (``NLP artifacts only''), and documentation doubled; (ii)~dropping each dimension in turn; and (iii)~1{,}000 \textit{random reweightings}, in which each of the six default weights is independently multiplied by a random factor between 0.5 and 1.5 (i.e.\ changed by up to $\pm$50\%).

\begin{table}[t]
\centering
\small
\setlength{\tabcolsep}{4pt}
\begin{tabular}{@{}lrccccc@{}}
\toprule
 & & \multicolumn{3}{c}{\textbf{Random}} & \multicolumn{2}{c}{\textbf{Equal}} \\
\cmidrule(lr){3-5}\cmidrule(l){6-7}
\textbf{Score} & $\boldsymbol{n}$ & $\boldsymbol{\rho}$ & \textbf{Same} & \textbf{Range} & $\boldsymbol{\rho}$ & \textbf{Same} \\
\midrule
0        & 3{,}609 & \multicolumn{5}{c}{\textit{unaffected by weighting}} \\
0--10    & 2{,}403 & 0.941 & 96.7 & 514 & 0.832 & 82.1 \\
10--20   & 518     & 0.866 & 77.1 & 244 & 0.220 & 79.3 \\
20--50   & 1{,}444 & 0.972 & 85.7 & 292 & 0.931 & 69.7 \\
50--100  & 194     & 0.992 & 96.6 & 17  & 0.950 & 94.3 \\
\bottomrule
\end{tabular}
\caption{Stability within each default score band under random reweighting (mean of 1{,}000 draws) and under equal weights. $\rho$: Spearman
correlation with the default ranking within the band. Same: percentage of
the band's languages that remain in the band. Range: median width of the
5th--95th percentile rank interval across the 1{,}000 draws.}
\label{tab:sensitivity-bands}
\end{table}

\begin{table}[t]
\centering
\small
\setlength{\tabcolsep}{2pt}
\begin{tabular}{@{}lccccc@{}}
\toprule
\textbf{Weighting} & $\boldsymbol{\rho}$ & \textbf{Top 20} & \textbf{Top 100} & \textbf{Shift} & \textbf{$<$10} \\
\midrule
Default               & 1.000 & 20   & 100  & 0     & 73.6 \\
Equal weights         & 0.959 & 17   & 91   & 211   & 68.5 \\
NLP artifacts only    & 0.931 & 19   & 96   & 283   & 75.8 \\
Documentation $\times$ 2 & 0.984 & 18   & 93   & 91    & 70.3 \\
\midrule
Without G             & 0.970 & 20   & 97   & 169.5 & 74.5 \\
Without D             & 0.951 & 19   & 98   & 239   & 74.3 \\
Without C             & 0.994 & 19   & 96   & 7.5   & 71.1 \\
Without M             & 0.892 & 20   & 95   & 299   & 77.4 \\
Without B             & 0.999 & 15   & 92   & 2     & 71.0 \\
Without P             & 0.980 & 19   & 94   & 59.5  & 72.3 \\
\midrule
Random (mean of 1k)     & 0.989 & 19.0 & 97.1 & 72    & 73.1 \\
\bottomrule
\end{tabular}
\caption{Sensitivity of the composite to the weighting, over the 4{,}559 active languages. $\rho$: Spearman correlation with the default ranking. Top 20 / Top 100: number of default top-ranked languages retained. Shift: median absolute change in rank. $<$10: percentage of all languages scoring
below 10.}
\label{tab:sensitivity}
\end{table}

The 3{,}609 zero-scoring languages (44.2\%) are not affected by any weighting. Of the 2{,}403 languages scoring above 0 but below 10, 96.7\% remain below 10 under $\pm 50\%$ perturbations, and 82.1\% under equal weights. The share of languages below 10 remains between 68.4\% and 75.9\% across draws (73.6\% by default). The relative order is largely preserved ($\rho = 0.941$), but individual
positions vary (median rank range 514). The 10--20 band is most sensitive to the weighting; under equal weights the order is mostly lost ($\rho = 0.220$). Above 50, ranks are nearly fixed ($\rho = 0.992$; median range 17). Overall, the ranking correlates with the default at a mean $\rho = 0.989$ (minimum 0.955) and is most sensitive to model coverage ($\rho = 0.892$ without it). Membership of the lowest band and the identity of the top languages are robust, while individual ranks below 50 should be read as indicative.

\section{Leave-One-Out Analysis}

Because some evaluated systems are also sources of the index, each of the 29 corpora, models, and benchmarks is removed and the composite score with default weights is recomputed.

\paragraph{Index Stability}For 28 of the 29 sources, the non-zero language ranking is largely unchanged (Spearman $\rho \geq 0.996$ overall; $\rho \geq 0.983$ among languages scoring below 10). The exception is Omnilingual, the only model for 499 languages, 452 of which have no other digital resource. Without it, these languages fall back to documentation-only scores ($\rho = 0.919$; $0.496$ below 10; 75.3\% below 10). The three broadest sources (FineWeb2, mmBERT, Apertus) leave the ranking intact ($\rho \geq 0.996$), but removing any one of them moves 435--771 languages from 20 or above to below 20. These sources determine the level of the middle range rather than the order within it.

\paragraph{Validation Stability}Removing any single source changes the correlation of the composite with task performance by at most 0.03 (MT $-0.029$ to $+0.017$; BR $-0.030$ to $+0.019$; NLI $-0.012$ to $+0.017$). Single dimensions are more sensitive; without MADLAD-400, the corpus correlation with MT falls from 0.49 to 0.33.

\paragraph{Removing Evaluated Systems}The validation is repeated after removing every source that is also an evaluated system or its training corpus: NLLB (corpus and model), Apertus, and MADLAD-400 for MT, XLM-R for BR, and XLM-R and mT5 for NLI. The composite is unaffected (MT 0.32 $\rightarrow$ 0.31; BR and NLI unchanged). Only the MT corpus correlation drops, because MADLAD-400 is the training data of the evaluated MADLAD model. XLM-R and mT5 cover languages already covered by broader models, so removing them changes no correlation. 

\begin{table}[!b]
\centering\small
\begin{tabular}{@{}llcccc@{}}
\toprule
\textbf{Task} & \textbf{Index} & \textbf{C} & \textbf{M} & \textbf{B} & \textbf{Comp.} \\
\midrule
MT  & full          & 0.49 & 0.37 & 0.48 & 0.32 \\
    & w/o evaluated & 0.33 & 0.37 & 0.48 & 0.31 \\
\midrule
BR  & full          & 0.31 & 0.27 & 0.47 & 0.23 \\
    & w/o evaluated & 0.31 & 0.27 & 0.47 & 0.23 \\
\midrule
NLI & full          & 0.27 & 0.21 & 0.36 & 0.23 \\
    & w/o evaluated & 0.27 & 0.21 & 0.36 & 0.23 \\
\bottomrule
\end{tabular}
\caption{Spearman correlation with task performance before and after removing
the evaluated systems from the index. G, D and P are unaffected.}
\label{tab:loo-eval}
\end{table}

\end{document}